%% file: main_icml.tex
\documentclass[]{article}

\usepackage{microtype}
\usepackage{graphicx}
\usepackage{subfigure}
\usepackage{booktabs} % for professional tables
\usepackage{hyperref}

\usepackage[accepted]{icml2022}

\usepackage{amsmath}
\usepackage{amssymb}
\usepackage{mathtools}
\usepackage{amsthm}
\usepackage{bbm}
\usepackage[]{cleveref}
\usepackage{comment}
\usepackage[textsize=tiny]{todonotes}
\usepackage[utf8]{inputenc}
\usepackage{microtype}
\usepackage{nicefrac}
\usepackage{physics}
\usepackage{subfigure}
\usepackage{xcolor}

\input{settings}
\newcommand{\textcite}[1]{\citet{#1}}

\icmltitlerunning{The dynamics of representation learning in shallow, non-linear autoencoders}

\begin{document}

\twocolumn[ \icmltitle{The dynamics of representation learning in shallow,
  non-linear autoencoders}

\begin{icmlauthorlist}
\icmlauthor{Maria Refinetti}{ens,epfl}
\icmlauthor{Sebastian Goldt}{sissa}
\end{icmlauthorlist}

\icmlaffiliation{ens}{Department of Physics, Ecole Normale Sup\'erieure, Paris,
  France}
\icmlaffiliation{epfl}{IdePHICS laboratory, EPFL, Lausanne, Switzerland}
\icmlaffiliation{sissa}{International School of Advanced Studies (SISSA), Trieste, Italy}

\icmlcorrespondingauthor{Sebastian Goldt}{sgoldt@sissa.it}

\icmlkeywords{Machine Learning, ICML}

\vskip 0.3in
]



\printAffiliationsAndNotice{~}  % leave blank if no need to mention equal contribution
\input{content/abstract}
\input{content/main_text}

\input{content/acknowledgements}
\bibliography{nn}
\bibliographystyle{icml2022}

\newpage
\appendix
\onecolumn
\numberwithin{equation}{section}

\input{content/appendix}

\end{document}

%% file: settings.tex
\newcommand{\calA}{\mathcal{A}}
\newcommand{\calB}{\mathcal{B}}
\newcommand{\calC}{\mathcal{C}}

\newcommand{\One}{\mathbbm{1}_{\left(\rho_\tau \in
    \mathopen[\rho,\rho+\varepsilon_\rho\mathclose[\right)}}

\definecolor{C0}{HTML}{1f77b4}
\definecolor{C1}{HTML}{ff7f0e}
\definecolor{C2}{HTML}{2ca02c}
\definecolor{C3}{HTML}{d62728}
\definecolor{C4}{HTML}{9467bd}
\definecolor{C5}{HTML}{8c564b}
\definecolor{C6}{HTML}{e377c2}
\definecolor{C7}{HTML}{7f7f7f}
\definecolor{C8}{HTML}{bcbd22}
\definecolor{C9}{HTML}{17becf}

\DeclareMathOperator{\pmse}{\sf pmse}

\newcommand{\sqD}{\sqrt{D}}

\newcommand{\usD}{\frac{1}{D}}
\newcommand{\usDD}{\frac{1}{D^2}}
\newcommand{\ussD}{\frac{1}{\sqrt{D}}}

\newcommand{\EE}{\mathbb{E}\,}

\newcommand{\reals}{\mathbb{R}}

\newcommand{\normal}{\mathcal{N}}

\newcommand{\bx}{\boldsymbol{x}}
\newcommand{\bw}{\boldsymbol{w}}
\newcommand{\bv}{\boldsymbol{v}}
\newcommand{\bc}{\boldsymbol{c}}
\newcommand{\bxi}{\boldsymbol{\xi}}
\newcommand{\balpha}{\boldsymbol{\alpha}}
\newcommand{\blambda}{\boldsymbol{\lambda}}
\newcommand{\bnu}{\boldsymbol{\nu}}
\newcommand{\bA}{\text{\textbf{A}}}
\newcommand{\bO}{\text{\textbf{O}}}
\newcommand{\bW}{\text{\textbf{W}}}
\newcommand{\bV}{\text{\textbf{V}}}
\newcommand{\bOmega}{\boldsymbol{\Omega}}
\newcommand{\T}{\text{\textbf{T}}}
\newcommand{\R}{\text{\textbf{R}}}
\newcommand{\Q}{\text{\textbf{Q}}}
\newcommand{\bulk}{\text{bulk}}
\newcommand{\equal}{\!=\!}

%% file: content/abstract.tex
% 4-6 sentences according to call for papers
\begin{abstract}
 Autoencoders are the simplest neural network for unsupervised learning, and thus an ideal framework for studying feature learning.
 While a detailed understanding of the dynamics of linear autoencoders has
 recently been obtained, 
 the study of non-linear autoencoders
 has been hindered by the technical difficulty of
 handling training data with non-trivial
 correlations -- a fundamental prerequisite for feature extraction.
 Here, we study the dynamics of feature learning in non-linear, shallow autoencoders.
 We derive a set of asymptotically exact equations that describe the
 generalisation dynamics of autoencoders trained with stochastic
 gradient descent (SGD) in the limit of high-dimensional inputs.
 These equations reveal that autoencoders learn the leading principal components of their inputs sequentially. 
 An analysis of the long-time dynamics explains the failure of sigmoidal autoencoders to learn with tied weights, and highlights the importance of training the bias in ReLU autoencoders.
 Building on previous results for linear networks, we analyse a modification of
 the vanilla SGD algorithm which allows learning of the exact principal components.
 Finally, we show that our equations accurately describe the generalisation dynamics  of non-linear autoencoders trained on realistic datasets such as~CIFAR10, thus establishing shallow autoencoders as an instance of the recently observed Gaussian universality.
\end{abstract}

%% file: content/main_text.tex
\section{Introduction}

Autoencoders are a class of neural networks trained to reconstruct their
inputs by minimising the distance between an input~$\bx\in\mathcal{X}$, and the network output~$\hat
\bx\in\mathcal{X}$. The key idea for learning good features with autoencoders is
to make the intermediate layer in the middle of the network (significantly) smaller than the
input dimension. This ``bottleneck'' forces the network to develop a compressed
representation of its inputs, together with an encoder and a
decoder. Autoencoders of this type are called undercomplete, see
\cref{fig:figure1}, and provide an ideal framework to study
feature learning, which appears to be a key property of deep learning, yet
remains poorly understood from a theoretical point of view.

How well can shallow autoencoders with a single hidden layer of neurons perform
on a reconstruction task? For \emph{linear} autoencoders,
\textcite{eckart1936approximation} established that the optimal reconstruction
error is obtained by a network whose weights are given by (a rotation of) the~$K$ leading principal components of the data, \emph{i.e.}~the $K$ leading
eigenvectors of the input covariance matrix. The reconstruction error of such a
network is therefore called the PCA error. How linear autoencoders learn these
components when trained using stochastic gradient descent (SGD) was first
studied by \textcite{bourlard1988auto} and \textcite{baldi1989neural}. A series
of recent work analysed linear autoencoders in more detail,
% since we move the related work to the discussion, be a bit more specific here
% and introduce main themes for the paper:
% convergence to the weights - i.e. SGD can find this optimal representation
% sequential learning
% we can also learn the actual eigenvectors by modifying the algorithm
focusing on the loss landscape~\cite{kunin2019loss}, the
convergence and implicit bias of SGD dynamics~\cite{pretorius2018learning,
  saxe2019mathematical, nguyen2019dynamics}, and on recovering the exact
principal components~\cite{gidel2019implicit, bao2020regularized,
  oftadeh2020eliminating}.

Adding a non-linearity to autoencoders is a key step on the way to more
realistic models of neural networks, which crucially rely on non-linearities in
their hidden layers to extract good features from data. On the other hand,
non-linearities pose a challenge for theoretical analysis. In the context of
high-dimensional models of learning, an additional complication arises from
analysing non-linear models trained on inputs with non-trivial correlations,
which are a fundamental prerequisite for unsupervised learning. The first step
in this direction was taken by \textcite{nguyen2021analysis}, who analysed
overcomplete autoencoders where the number of hidden neurons grows
polynomially with the input dimension, focusing on the special case of
weight-tied autoencoders, where encoder and decoder have the same weights.

Here, we leverage recent developments in the analysis of supervised learning on
complex input distributions to study the learning dynamics of undercomplete,
non-linear autoencoders with both tied and untied weights. While their
reconstruction error is also limited by the PCA reconstruction
error~\cite{bourlard1988auto, baldi1989neural}, we will see that their learning
dynamics is rich and raises a series of questions: can non-linear AE trained
with SGD reach the PCA error, and if so, how do they do it? How do the
representations they learn depend on the architecture of the network, or the
learning rule used to train them?\\

\noindent We describe our setup in detail in \cref{sec:setup}. Our main contributions can
be summarised as follows:

\begin{enumerate}
\item We derive a set of asymptotically exact equations that describe the
  generalisation dynamics of shallow, non-linear autoencoders trained using
  online stochastic gradient descent (SGD) (\cref{sec:eom});
\item Using these equations, we show how autoencoders learn important features
  of their data sequentially (\ref{sec:sequential-learning}); we analyse the
  long-time dynamics of learning and highlight the importance of untying the
  weights of decoder and encoder (\ref{sec:long-time}); and we show that
  training the biases is necessary for ReLU autoencoders;
  (\ref{sec:importance-bias-relu})
\item We illustrate how a modification of vanilla SGD breaking
  the rotational symmetry of neurons yields the exact principal components of
  the data (\cref{sec:truncated-sgd});
\item We finally show that the equations also capture the learning dynamics of
  non-linear autoencoders trained on more realistic datasets such as CIFAR10
  with great accuracy, and discuss connections with recent results on Gaussian
  universality in neural network models in \cref{sec:realistic-data}.
\end{enumerate}

% The paper is organised as follows.
% We describe our setup in detail in \cref{sec:setup} before sketching the
% derivation of the equations describing the learning dynamics in \cref{sec:eom}. Our analysis 
% In \cref{sec:truncated-sgd}, we analyse a modification of SGD which allows to retrieve the eigenvectors.
% We tackle the long time dynamics of non-linear AE in \cref{sec:long-time},
% where we uncover the importance of untying the weights and of training the bias
% in sigmoidal and  ReLU autoencoders, respectively.
% Finally, we demonstrate that our theoretical results carry over to learning on
% realistic datasets in \cref{sec:realistic-data} and conclude with a discussion
% of our results in relation to further previous work in \cref{sec:discussion}.

\paragraph{Reproducibility} We provide code to solve the dynamical equations of \cref{sec:eom} 
and to reproduce our plots at
\href{https://github.com/mariaref/NonLinearShallowAE}{https://github.com/mariaref/NonLinearShallowAE}.

\section{Setup}
\label{sec:setup}

\begin{figure}[t!]
  \centering
  \includegraphics[width=\linewidth]{{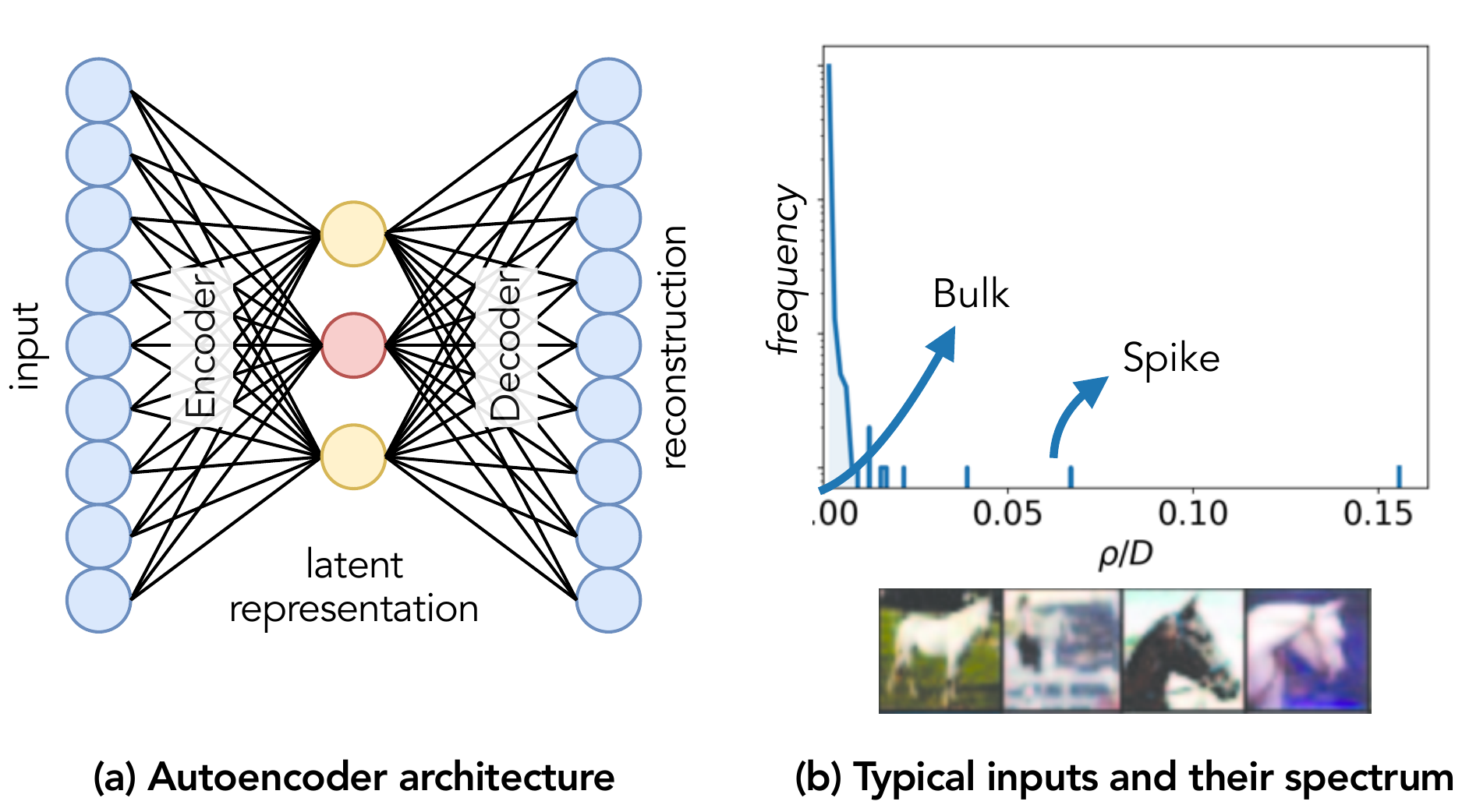}}
  \caption{\label{fig:figure1} \textbf{Shallow autoencoders and their inputs}
    \textbf{(a)} We analyse two-layer autoencoders with non-linear activation
    function in the hidden layer. %
    \textbf{(b)} Top: Rescaled eigenvalues of the covariance matrix of inputs
    drawn from the spiked Wishart model in \cref{eq:dataset}, which can be
    divided into a bulk and a finite number of outliers. Bottom: example inputs
    drawn from CIFAR10~\cite{krizhevsky2009learning}, a benchmark data set we
    use for our experiments with realistic data in \cref{sec:realistic-data}.}
\end{figure}

\paragraph{Architecture} We study the performance and learning dynamics of
shallow non-linear autoencoders with $K$ neurons in the hidden layer. Given a
$D$-dimensional input $\bx = (x_i)$, the output of the autoencoder is given by
\begin{equation}
  \label{eq:ae}
  \hat x_i = \sum_k^K v_i^k g\left( \lambda^k\right), \qquad \lambda^k \equiv \frac{\bw^k \bx}{\sqrt D}
\end{equation}
where $\bw^k, \bv^k \in\reals^D$ are the encoder and decoder
weight of the $k$th hidden neuron, resp., and $g(\cdot)$ is a
non-linear function. We study this model 
in the thermodynamic limit where we let the input dimension~$D\!\to\!\infty$ while
keeping the number of hidden neurons $K$ finite, as shown in \cref{fig:figure1}
(a). The performance of a \emph{given}
autoencoder is measured by the population reconstruction mean-squared error,
\begin{equation}
    \label{eq:pmse}
    \pmse \equiv \frac{1}{D}\sum_i\mathbb E  (x_i-\hat x_i)^2,
\end{equation}
where the expectation $\EE$ is taken with respect to the data
distribution. \textcite{nguyen2021analysis} analysed shallow autoencoders in a
complementary mean-field limit, where the number of hidden neurons $K$ grows
polynomially with the input dimension. Here, by focusing on the case $K\!<\!D$, we
study the case where the hidden layer is a bottleneck.
\textcite{nguyen2021analysis} considers tied autoencoders, where a single set of
weights is shared between encoder and decoder, $w^k\equal v^k$.  We will see in
\cref{sec:long-time} that untying the weights is critical to achieve a
non-trivial reconstruction error in the regime~$K<D$.

\paragraph{Data model} We derive our theoretical results for inputs drawn from a spiked
Wishart model~\cite{wishart1928generalised, potters2020first}, which has been
widely studied in statistics to analyse the performance of unsupervised learning
algorithms. In this model, inputs are sampled according to
\begin{equation}
    \label{eq:dataset}
    \bx_\mu = \bA \bc_\mu + \sqrt \sigma \bxi_\mu,
\end{equation}
where $\mu$ is an index that runs over the inputs in the training set,
$\bxi_\mu$ is a $D$-dimensional vector whose elements are drawn i.i.d.~from a
standard Gaussian distribution and~$\sigma\!>\!0$. The matrix $\bA\!\in\!\mathbb{R}^{D
  \times M}$ has elements of order $\order{1}$ and is fixed for all inputs, while we sample~$\bc_\mu\!\in\!\mathbb{R}^{M}$ from some distribution. Different choices for~$\bA$ and the distribution of $\bc$ allow modelling of Gaussian mixtures, sparse codes, and non-negative sparse coding.

\paragraph{Spectral properties of the inputs} The spectrum of the covariance of
the inputs is determined by~$\bA$ and $\bc$. It governs the dynamics and the
performance of autoencoders (by construction, we have~$\EE x_i\!=\!0$). If we
set~$\bc_\mu\!=\!0$ in \cref{eq:dataset}, inputs would be i.i.d.~Gaussian
vectors with an average covariance of $\mathbb{E} x_i x_j\!=\!\delta_{ij}$, where
$\delta_{ij}$ is the Kronecker delta. The \emph{empirical} covariance matrix of
a finite data set with~$P\!\sim\!\order{D}$ samples has a Marchenko–Pastur
distribution with a scale controlled
by~$\sigma$~\cite{marchenko1967distribution, potters2020first}. By sampling the
$m$th element of $\bc$ as $c_m\!\sim\!\normal(0, \tilde{\rho}_m)$, we ensure that the
input-input covariance has $M$ eigenvalues $\rho_m\equal D\tilde{\rho}_m$ with the columns of~$\bA$ as
the corresponding eigenvectors. The remaining $D\!-\!M$ eigenvalues of the empirical
covariance, i.e.~the \emph{bulk}, still follow the Marchenko-Pastur
distribution. We further ensure that the reconstruction task for the autoencoder
is well-posed by letting the outlier eigenvalues scale
as~$\rho_m\!\sim\!\order{D}$ (or $\tilde{\rho}_m
\!\sim\!\order{1}$), resulting in spectra such as the one shown in
\cref{fig:figure1}(b). 
If the largest eigenvalues were of order one, an
autoencoder with a finite number of neurons~$K$ could not obtain a $\pmse$
better than random guessing as the input dimension~$D\!\to\!\infty$.

\paragraph{Training} We train the autoencoder on the quadratic error in the
one-pass (or online) limit of stochastic gradient descent (SGD), where a
previously unseen input $\bx$ is presented to the network at each step of
SGD. This limit is extensively used to analyse the dynamics of non-linear
networks for both supervised and unsupervised learning, and it has been shown
experimentally to capture the dynamics of deep networks trained with data
augmentation well~\cite{nakkiran2020bootstrap}.  The SGD weight increments for
the network parameters are then given by:
\begin{subequations}
  \label{eq:sgd}
  \begin{align}
  % \dd w_{i}^k &\equiv \left(w_{i}^k\right)_{\mu+1}- \left(w_{i}^k\right)_{\mu}
% \\
    \dd w_{i}^k&=-\frac{\eta_{W}}{D}\left( \frac{1}{\sqD}\sum_j v_j^k \Delta_j g'(\lambda^k) x_i\right) - \frac{\kappa}{D} w_i^k, \label{eq:sgdw}\\
  \dd v_i^k &= - \frac{\eta_{V}}{D} g(\lambda^k) \Delta_i - \frac{\kappa}{D} v_i^k, \label{eq:sgdv}
  \end{align}
\end{subequations}
where $\Delta_i \equiv (\hat x_i - x_i)$ and $\kappa\!\geq\! 0$ is the $\ell_2$ regularisation constant.  In order to obtain a well
defined limit in the limit $D\!\to\!\infty$, we rescale the learning rates as $\eta_W\equal \eta / D$, $\eta_V\equal \eta$ for some fixed~$\eta\!>\! 0$.

\section{Results}

\subsection{Dynamical equations to describe feature learning}
\label{sec:eom}

The starting point for deriving the dynamical equations is the observation that
the $\pmse$ defined in \cref{eq:pmse} depends on the inputs only via their
low-dimensional projections on the network's weights
\begin{equation}
  \label{eq:local-fields}
  \lambda^k \equiv \frac{\bw^k \bx}{\sqrt D}, \qquad \nu^k \equiv \frac{\bv^k \bx}{D}.
\end{equation}
This allows us to replace the
high-dimensional expectation over the inputs in \cref{eq:pmse} with a
$K$-dimensional expectation over the local fields $\blambda\equal (\lambda^k)$ and
$\bnu\equal (\nu^k)$. For Gaussian inputs drawn from \eqref{eq:dataset}, $(\blambda, \bnu)$ are jointly Gaussian and their distribution is entirely characterised by their
second moments:
\begin{align}
    \label{eq:order_parameters}
    T_1^{k \ell} \equiv \EE \nu^k \nu^\ell & = \usDD \sum_{i,j} v^k_i \Omega_{ij} v^\ell_j, \\
    \quad Q_1^{k \ell} \equiv \EE \lambda^k \lambda^\ell & = \usD \sum_{i,j}w^k_i \Omega_{ij} w^\ell_j,\\ \quad
  R_1^{k \ell} \equiv \EE \nu^k \lambda^\ell & = \frac{1}{D^{\nicefrac{3}{2}}} \sum_{i,j} v^k_i \Omega_{ij} w^\ell_j.
\end{align}
where $\Omega_{ij} = \EE x_i x_j$ is the covariance of the inputs. These macroscopic overlaps, often referred to as \emph{order
  parameters} in the statistical physics of learning, together with
$T^{kl}_0\!~=~\! \sum_i\nicefrac{v_i^k v_i^\ell}{D}$,  are sufficient to evaluate
the $\pmse$. To analyse the dynamics of learning, the goal is then to derive a
closed set of differential equations which describe how the order parameters
evolve as the network is trained using SGD~\eqref{eq:sgd}. Solving these
equations then yields performance of the network at all times. Note that feature
learning requires the weights of the encoder and the decoder move far away from
their initial values.  Hence, we cannot resort to the framework of the neural
tangent kernel or \emph{lazy learning}~\cite{jacot2018neural, chizat2019lazy}.

Instead, here we build on an approach that was pioneered by \textcite{Saad1995a,
  Saad1995b} and \textcite{Riegler1995} to analyse supervised learning in
two-layer networks with uncorrelated inputs; \textcite{saad2009line} provides a
summary of the ensuing work. We face two new challenges when analysing autoencoders in this setup. First, autoencoders cannot be formulated in the usual teacher-student setup, where a student network is trained on inputs whose label is provided by a fixed teacher network, because it is impossible to choose a teacher autoencoder which exactly implements the ``identify function'' $x\to x$ with $K<D$. Likewise, a teacher autoencoder with random weights, the typical choice in supervised learning, is not an option since such an autoencoder does not implement the identity function, either. Instead, we bypass introducing a teacher and develop a description starting from the data properties. The second challenge is then to handle the non-trivial correlations in the inputs~\eqref{eq:dataset}, where we build on recent advances in supervised learning on structured data~\cite{yoshida2019datadependence, goldt2020modelling,
  goldt2021gaussian, refinettiClassifying}.

\subsubsection{Derivation: a sketch}

Here, we sketch the derivation of the equation of motion for the order parameter
$\T_1$; a complete derivation for $\T_1$ and all other order parameters is given in \cref{sec:odes}.
We define the eigen-decomposition of the covariance matrix
$\Omega_{rs}\equal \nicefrac{1}{D} \sum_{\tau=1}^D \Gamma_{s\tau} \Gamma_{r\tau}
\rho_\tau$ and the rotation of any vector $z \in\{w^k, v^k, x\}$ onto this basis
as $z_\tau\!\equiv\! \nicefrac{1}{\sqrt D}\sum_{s=1}^D \Gamma_{\tau s} z_s$. The eigenvectors are normalised as $\sum_{i} \Gamma_{\tau i} \Gamma_{\tau' i}\equal  D \delta_{\tau \tau'}$ and $\sum_{\tau} \Gamma_{\tau i} \Gamma_{\tau j}\equal  D \delta_{ij}$. Using this decomposition, we can re-write $T_1$ and its update at step $\mu$ as: 
\begin{align}
% T_1^{k \ell} &= \usDD \sum_\tau \rho_\tau v_\tau^k v_\tau^\ell,\\
\left(T_1^{k \ell}\right)_{\mu\!+\!1}\!\!\!\!-\left(T_1^{k \ell}\right)_{\mu}\!\!&=\!\usDD\!\sum_\tau\!\rho_\tau\!\left(\dd v_\tau^k v_\tau^\ell \!+ \! v_\tau^k \dd v_\tau^\ell \! +\!\dd v_\tau^k \dd v_\tau^\ell \right).\notag
\end{align}
% Naively replacing the SGD update Eq.~\ref{eq:sgd} into the second equation leads to expressions which cannot be simply expressed in terms of other order parameters. 
We introduce the order-parameter density $t(\rho,s)$ that depends on $\rho$ and
on the normalised number of steps $s\equal \nicefrac{\mu}{D}$, which we interpret as a
continuous time variable in the limit~$D\!\to\!\infty$,
\begin{align}
  \label{eq:ttau}
  t^{k\ell}(\rho,s)&=\frac{1}{D^2 \epsilon_\rho}  \sum_\tau v^k_\tau v^\ell_\tau \One,
\end{align}
where $\mathbbm{1}(.)$ is the indicator function and the limit $\epsilon_\rho\!\to\! 0$
is taken after the thermodynamic limit $D\to\infty$. Similarly $r(\rho,s)$ and $q(\rho,s)$ are introduced as order parameter densities corresponding to $R_1$ and $Q_1$ respectively. 
Now, inserting the update for $\bv$~\eqref{eq:sgd} in the expression of $t^{k\ell}$, and evaluating the expectation over a fresh sample $\bx$, we obtain the dynamical equation:
\begin{multline}
\frac{\partial t^{k\ell}(\rho, s) }{\partial s}
 = \eta \left( \frac{\EE \lambda^k g(\lambda^k)}{Q_1^{kk}} \frac{\rho}{D} r^{\ell k}(\rho, s)\right.\\
 \left. \qquad \quad - \sum_a t^{\ell k}(\rho, s)  \EE g(\lambda^k) g(\lambda^a)
  \right)  + (k\leftrightarrow \ell ).
   \label{eq:evolution_t_tau}
\end{multline}
where we write $k\leftrightarrow \ell$ to denote the same term with the indices $k$ and $\ell$ permuted. The order parameter is recovered by integrating the density~$t^{\ell k}(\rho, s)$ over the
spectral density $\mu_\Omega$ of the covariance matrix: $T_1^{k\ell}\!\!~=~\!\!\int\!\! \dd \mu_\Omega(\rho)
\;  \nicefrac{\rho}{D} \; t^{k\ell}(\rho,s)$.

We can finally close the equations by evaluating the remaining averages such as
$\EE \lambda^k g(\lambda^k)$. For some activation functions, such as the ReLU or
sigmoidal activation~$g(x)\equal \erf\left( \nicefrac{x}{\sqrt 2} \right)$,
these integrals have an analytical closed form. In any case, the averages only
depend on the second moments of the Gaussian random variables $(\blambda, \bnu)$,
which are given by the order parameters, allowing us to close the equations on
the order parameters.

We derived these equations using heuristic methods from statistical physics. While we conjecture that they are exact, we leave it to future work to prove their asymptotic correctness in the limit $D\to\infty$. We anticipate that techniques to prove the correctness of dynamical equations for supervised learning \citet{goldt2019dynamics, veiga2022phase} should extend to this case after controlling for the spectral dependence of order parameters.

\subsubsection{Separation between bulk modes and principal components} 

We can exploit the separation of bulk and outlier eigenvalues to decompose the
integral~\eqref{eq:ttau} into an integral over the \emph{bulk} of the spectrum
and a sum over the outliers. This results in a decomposition of the overlap $\T_1$ as:
\begin{align}
    \T_1 &= \sum_{i\in\text{outliers}}^M \tilde{\rho}_i t_i + \T_{\bulk}, \\
    \T_{\bulk} &=\frac 1 D \int_{\rho\in {\bulk}}\dd \mu_\Omega(\rho)\rho t(\rho),
\end{align}
and likewise for other order parameters. Leveraging the fact that the bulk
eigenvalues are of order $\order{1}$, and that we
take the $D\to\infty$ limit, we can write the equation for $\T_{\bulk}$ as:
\begin{align}
   \frac{\partial T^{kl}_{\bulk} }{\partial s}
  &= -\eta \sum_a  T^{al}_{\bulk} \EE g(\lambda^k) g(\lambda^a) + (k\leftrightarrow \ell ).
\end{align}
On the other hand, the $M$ outlier eigenvalues, i.e.~the spikes, are of order
$\order{D}$. We thus need to keep track of all the terms in their equations of
motion. Similarly to \cref{eq:evolution_t_tau}, they are written:
% \begin{multline}
%  \label{eq:evolution_t_tau}
% \frac{\partial t^{k\ell}_\tau }{\partial s}
% = \eta \left( \frac{\EE \lambda^k g(\lambda^k)}{Q_1^{kk}} \tilde{\rho}_\tau r^{\ell k}_\tau\right. \\
%  \left. - \sum_a t^{\ell k}_\tau  \EE g(\lambda^k) g(\lambda^a)
%  \right)  + (k\leftrightarrow \ell )
% \end{multline}
\begin{align}
 \label{eq:evolution_t_spike}
\frac{\partial t^{k\ell}_\tau }{\partial s}
= &\eta \left( \frac{\EE \lambda^k g(\lambda^k)}{Q_1^{kk}} \tilde{\rho}_\tau r^{\ell k}_\tau - \sum_a t^{\ell k}_\tau  \EE g(\lambda^k) g(\lambda^a)
 \right)\notag\\
 &+ (k\leftrightarrow \ell )
\end{align}
Note that the evolution of $\{t_i\}_{i=1,..,M}$ is coupled to the one of $\T_{\bulk}$ only through the expectations of the local fields. A similar derivation, which we defer to \cref{app:derivation_odes}, can be carried out for $\R_1$ and $\Q_1$.

\begin{figure}[t!]
  \centering
  \includegraphics[width=0.8\linewidth]{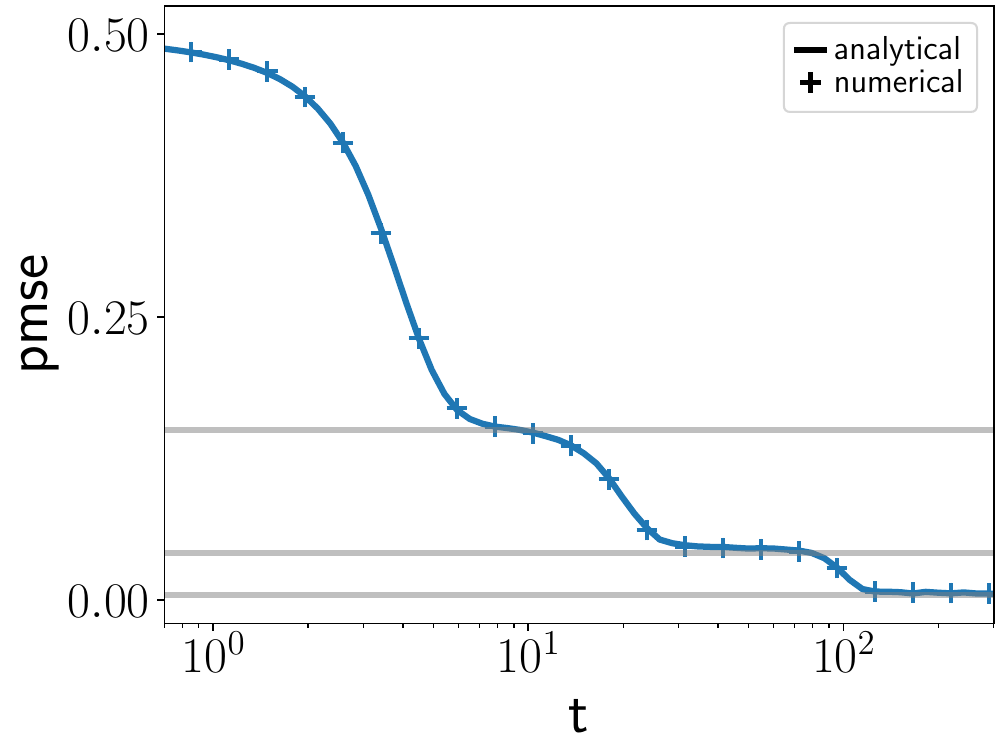}
  \caption{\label{fig:verification} \textbf{The dynamical equations describe the
      learning dynamics of autoencoders} Prediction mean-squared error
    $\pmse$~\eqref{eq:pmse} of an autoencoder with $K\equal 3$ hidden neurons during
    training on with stochastic gradient descent on the spiked Wishart model in
    the one-pass limit. We obtained the $\pmse$ directly from a simulation
    (crosses) and from integration of the dynamical equations describing
    learning that we derive in \cref{sec:eom} (line). Horizontal lines indicate
    the PCA reconstruction error with increasing numbers of principal components. \emph{Parameters:} $M\equal 3, \eta\equal 1$, $D\equal 1000$.}%, $\kappa\equal 0$.}
\end{figure}

\subsubsection{Empirical verification} In \cref{fig:verification}, we plot the
$\pmse$ of an autoencoder with $K\equal 3$ hidden neurons during training on a dataset
drawn from \cref{eq:dataset} with $M\equal 3$. We plot the $\pmse$ both obtained from
training the network using SGD~\eqref{eq:sgd} (crosses) and obtained from
integrating the equations of motion that we just derived (lines). The agreement
between the theoretical description in terms of bulk and spike modes captures
the training dynamics well, even at an intermediate input dimension of
$D\equal 1000$. In the following, we analyse these equations, and hence the behaviour of the autoencoder during training, in more detail.

\begin{figure*}[ht]
  \centering
  \includegraphics[width=\linewidth]{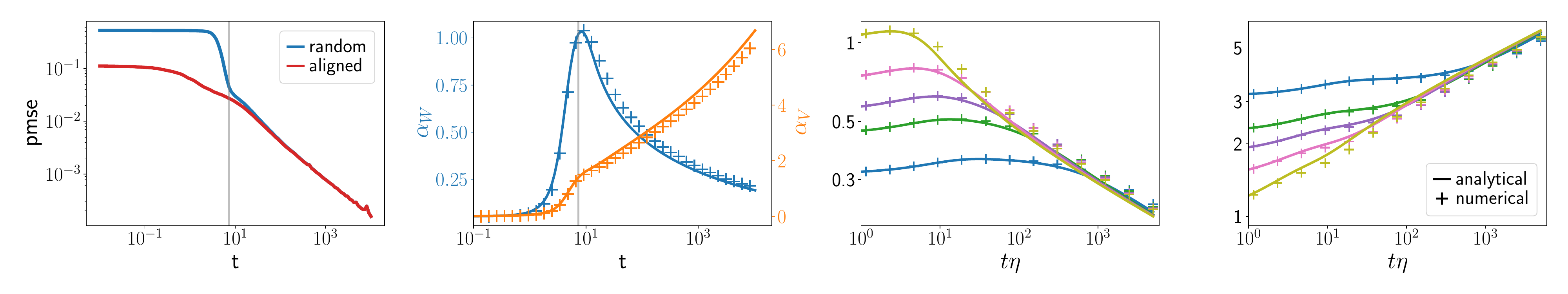}
  \vspace*{-2em}
  \caption{\label{fig:longtimedynamics} \textbf{Two phases of learning in
      sigmoidal autoencoders.} \textbf{(a)} Prediction error
    $\pmse$~\eqref{eq:pmse} of a sigmoidal autoencoder with a single hidden
    neuron starting from random initial conditions (blue). The error first
    decays exponentially, then as a power law. During this second phase of
    learning, the $\pmse$ is identical to the $\pmse$ of an autoencoder whose
    weights are proportional to the leading eigenvector $\Gamma$ of the input
    covariance at all times:
    $w(t) \propto \alpha_w(t) \Gamma, \quad v(t) \propto \alpha_v(t)\Gamma$,
    cf.~\cref{eq:ansatz_longtimes}. \textbf{(b)} Dynamics of the scaling
    constants $\alpha_w$ and $\alpha_v$ obtained through simulations of an
    autoencoder starting from random weights (crosses) and from integration of a
    reduced set of dynamical equations, \cref{eq:reduced-eom}. \textbf{(c,d)}
    The norm of the encoder (left) and decoder (right) weights. Their weights
    shrink, respectively grow, as a power law $\sqrt{\rho} (\eta t)^{\pm\delta}$
    with exponent $\delta\simeq 1/6$. This allows the sigmoidal network to
    remain in the linear region of its activation function and to achieve PCA
    performance. Solid lines are obtained from integration of a reduced set of
    dynamical equations, \cref{eq:reduced-eom}, crosses are from simulations for
    various combinations of learning rates and leading eigenvalue $\rho$
    (different colours). \emph{Parameters:}~$D=500, \eta=1 (a,b), M=1, K=1$.}
\end{figure*}

\subsection{Nonlinear autoencoders learn principal components sequentially}
\label{sec:sequential-learning}
%\begin{itemize}
%    \item \todo{In this section, discuss two points}:
%    \item How the bulk order parameters decay to zero.
%    \item How the eigenvalues rescale the learning rates of the %different modes, leading to a sequential learning of the %eigenmodes
%\end{itemize}

The dynamical equations of the encoder-encoder overlap $q^{k\ell}$, \cref{eq:q_tau_ode}, takes the form % spike mode order parameters~\eqref{eq:t_tau_ode},
%and~\eqref{eq:r_tau_ode} take the form:
\begin{align}
    \frac{\partial q^{k\ell}_1(\rho, t)}{\partial t} &= \frac{\eta\rho}{D}\left\{\cdots\right\}.
\end{align}
The learning rate of each mode of~$q^{k\ell}$ is thus rescaled by the corresponding eigenvalue, suggesting that the principal components that correspond to each to each mode are learnt at different speeds, with the modes corresponding to the largest eigenvalues being learnt the fastest. 
For the other order parameters $r^{k \ell}$ and $t^{k \ell}$, all the terms in the equations of motion \cref{eq:t_tau_ode} and~\cref{eq:r_tau_ode} are rescaled by $\rho$ except for one: 
\begin{align}
%    \begin{split}
        \frac{\partial r^{k\ell}_1(\rho, t)}{\partial t} &= \frac{\eta\rho}{D}\left\{\cdots\right\} - \eta\sum_a r^{ak}(\rho, s) I^{\ell a}_2(Q_1) ,\\
        \frac{\partial t^{k\ell}_1(\rho, t)}{\partial t} &= \frac{\eta\rho}{D}\left\{\cdots\right\} -\eta\sum_a t^{\ell a}(\rho, s) I^{ka}_2(Q_1)
%    \end{split}
\end{align}
%{\color{C1} Option 2:
%\begin{align}
%   \frac{\partial q_1(\rho, t)}{\partial t} &= \frac{\eta\rho}{D}\left\{\cdots\right\},\notag\\
%    \frac{\partial r_1(\rho, t)}{\partial t} &= \frac{\eta\rho}{D}\left\{\cdots\right\} - \eta I_2(Q_1) r(\rho, s),\\
%    \frac{\partial t_1(\rho, t)}{\partial t} &= \frac{\eta\rho}{D}\left\{\cdots\right\} -\eta I_2(Q_1) t(\rho, s) \notag
%\end{align}
% }
Indeed, the $\pmse$ of the sigmoidal autoencoder shown in \cref{fig:verification}
goes through several sudden decreases, preceded by plateaus with quasi-stationary $\pmse$.
Each transition is related to the network ``picking up'' an additional principal
component. By comparing the error of the network on the plateaus to the PCA
reconstruction error with an increasing number of principal components (solid
horizontal lines in \cref{fig:verification}), we confirm that the plateaus
correspond to stationary points where the network has learnt (a rotation of) the
first leading principal components. Whether or not the plateaus are clearly
visible in the curves depends on the separation between the eigenvalues of the leading eigenvectors.

This sequential learning of principal components has also been observed in
several other models of learning. It appears in unsupervised learning rules such
as Sanger's rule~\cite{sanger1989optimal} (aka as generalised Hebbian learning, cf. \cref{app:online-pca})
that was analysed by \textcite{biehl1998dynamics}% , and in supervised models of
% online learning in linear~\cite{saxe2014exact, advani2020highdimensional} and
% non-linear networks~\cite{Schwarze1993a, Saad1995a, Saad1995b, Riegler1995}
. \textcite{Gunasekar2017, gidel2019implicit} found a sudden
transition from the initial to the final error in linear models $\hat \bx = \bW \bx$,
but step-wise transitions for factorised models, i.e.~linear autoencoders of the
form $\hat \bx = \bV \bW \bx$. \textcite{saxe2019mathematical} also highlighted
the sequential learning of principal components, sorted by singular values.

The evolution of the bulk order parameters obeys
\begin{align}
    \begin{split}
         \frac{\partial T^{kl}_{\bulk} }{\partial s}
&= -\eta \sum_a  T^{al}_{\bulk} \EE g(\lambda^k) g(\lambda^a) + (k\leftrightarrow \ell )\\
 \frac{\partial R^{kl}_{\bulk} }{\partial s}
&= -\eta \sum_a  R^{al}_{\bulk} \EE g(\lambda^k) g(\lambda^a), \quad
 \frac{\partial Q^{kl}_{\bulk} }{\partial s} = 0
    \end{split}
\end{align}

As a consequence, in the early stages of training, to first approximation, the bulk component of $\R_1$ and $\T_1$ follow an exponential decay towards $0$. The characteristic time of this decay is given by the expectation $\EE g(\lambda^k)g(\lambda^\ell)\vert_{s=0}$ which only depends on $\Q_1\vert_{s=0}$. In addition, the last equation implies that the evolution of $\Q_1$ is entirely determined by the dynamics of the spike modes. 

%\begin{itemize}
%    \item \todo{Can we relate these plateaus to fixed points of the equations, or will the equations be singular for the neurons that have recovered the principal components?}
%\end{itemize}

\subsection{The long-time dynamics of learning: align, then rescale}
\label{sec:long-time}

%\begin{itemize}
%    \item say that we can study the long time dynamics : ansatz on the weights -> ansatz on eigenvecotrs -> ODES for scaling constants
%    \item ODEs describe well training 
%    \item Show how sigmoidal networks learn in two steps: first align to (a rotation of) the eigenvectors, Fig.~\ref{fig:sigmoidal} (a).
%    \item Then rescale the weights 
%    \item scaling law  -> rescale decoder and encoder to become as "linear" as possible, Fig.~\ref{fig:sigmoidal} (b).
%   \item prediction that a tied weight cannot learn --> confirmation by plot
%\end{itemize}

At long training
times, we expect a sigmoidal autoencoder with untied weights to have retrieved
the leading PCA subspace since it achieves the PCA reconstruction error. This
motivates an ansatz for the dynamical equations where the network weights are
proportional to the eigenvectors of the covariance~$\Gamma^k$, 
\begin{align}
  \label{eq:ansatz_longtimes}
    w_i^k(t) = \nicefrac{\alpha^k_\omega(t)}{\sqD} \; \Gamma_i^k \qquad v_i^k = \alpha^k_v(t)\Gamma_i^k,
\end{align}
where $\balpha_\omega, \balpha_v\in\reals^K$ are scaling
constants that evolve in time. With this ansatz, the order parameters are diagonal, 
% \begin{align}
% \begin{split}
% &Q_1^{k\ell}= \alpha^{k2}_w \tilde\rho_k \delta^{k\ell},\quad
%     R_1^{k\ell}= \alpha^{k}_w  \alpha^k_v \tilde\rho_k \delta^{k\ell},\\
%     &
%     T_1^{k\ell}= \alpha^{k2}_v \tilde\rho_k \delta^{k\ell},\quad
%     T_0^{k\ell}= \alpha^{k2}_v \delta^{k\ell}.
% \end{split}
% \end{align}
and we are left with a reduced set of $2 K$ equations describing the dynamics of
$\alpha_w^k$ and $\alpha^k_v$ which are given in \cref{eq:reduced-eom} together
with their detailed derivation.

\begin{figure*}[ht]
  \centering
  \includegraphics[width=\linewidth]{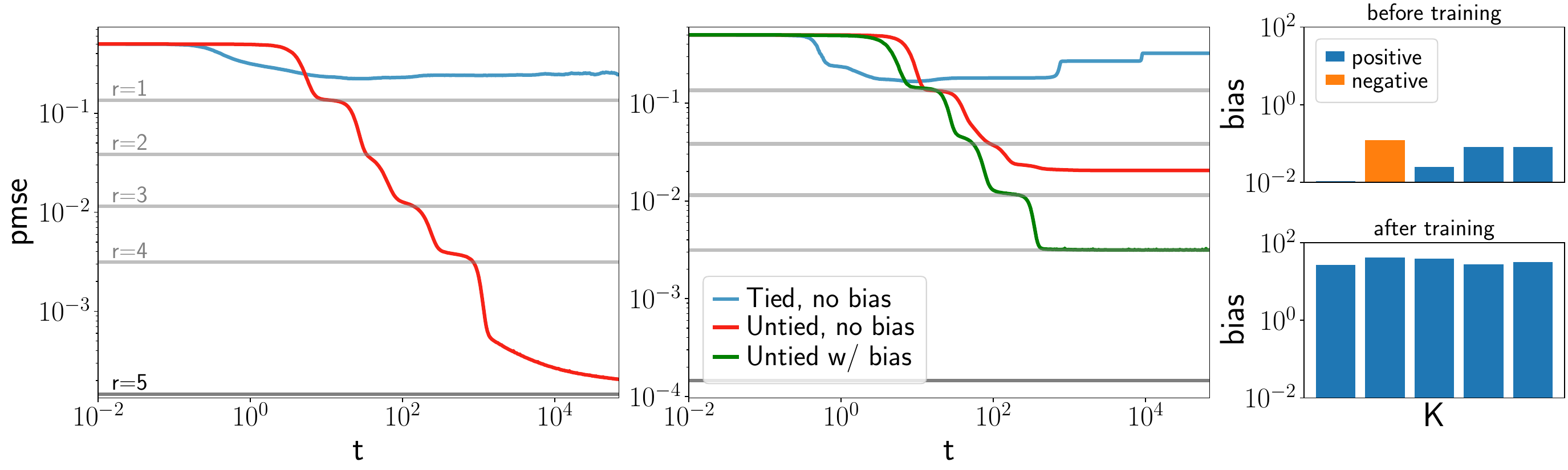}
  \caption{\label{fig:tiedvsuntied_synthetic}\textbf{Shallow autoencoders
      require untied weights, and ReLU autoencoders also need biases.}
    \textbf{(a)} Prediction error $\pmse$~\eqref{eq:pmse} of sigmoidal
    autoencoders with untied weights (red) and with tied weights $v^k=w^k$
    (blue). Horizontal lines indicate PCA errors for rank $r$. \textbf{(b)} Same
    plot for ReLU autoencoders with three different architectures: tied weights,
    no bias (blue), untied weights without bias (red) and untied weights with
    bias (green). Only the latter architecture achieves close to PCA
    error. \textbf{(c)} Distribution of biases in a ReLU network before (top)
    and after training (bottom). The biases increase throughout training, hence
    pushing the network into the linear region of its activation
    function. \emph{Parameters:} $D= 1000, K=5, \eta = 1$.}
\end{figure*}

We first verify the validity of the reduced equations~\eqref{eq:reduced-eom}
to describe the long-time dynamics of sigmoidal autoencoders. In
\cref{fig:longtimedynamics}(a), we show the generalisation dynamics of a
sigmoidal autoencoder starting from random initial conditions (blue). We see two
phases of learning: after an exponential decay of the $\pmse$ up to time
$t \sim 10$, the $\pmse$ decays as a power-law. This latter decay is well
captured by the evolution of the $\pmse$ of a model which we initialised with
weights aligned to the leading PCs. We deduce that during the exponential decay
of the error, the weights align to the leading PCA directions and stay aligned
during the power-law decay of the error.

After
having recovered the suitable subspace, the network adjusts the norm of its
weights to decrease the error. A look at the dynamics of the scale parameters
$\balpha_w$ and $\balpha_v$ during this second phase of learning reveals that the
encoder's weights shrink, while the decoder's weights grow,
cf.~\cref{fig:longtimedynamics}(b). Together, these changes lead to the
power-law decay of the $\pmse$. Note that we chose a dataset with only one
leading eigenvalue so as to guarantee that the AE recovers the leading principal
component, rather than a rotation of them. A scaling analysis of the evolution of the scaling constants $\alpha^k_w$ and $\alpha^k_v$ shows that the weights decay,
respectively grow, as a power-law with time,
$\alpha^k_w\propto \nicefrac{1}{\sqrt{\rho}}\; (\eta t)^{-\delta}$ and
$\alpha^k_v\propto \nicefrac{1}{\sqrt{\rho}}\; (\eta
t)^{\delta}$, see \cref{fig:longtimedynamics}(c-d). 

We can understand this behaviour by recalling the linearity of
$g(x)~\!=~\!\erf\left(\nicefrac{x}{\sqrt 2}\right)$ around the origin, where
$g(x)~\!\sim\!~\sqrt{\nicefrac{2}{\pi}} \; x$.
% \begin{equation}
%     g\left(\frac{x w^k}{\sqrt{D}}\right) \sim g\left(0 \right) + \frac{x w^k}{\sqrt{D}} g'(\lambda^k).
% \end{equation}
% If the encoder's weights $\bw$ are sufficiently small, higher order terms can be neglected. 
By shrinking the encoder's weights, the autoencoder thus performs a linear
transformation of its inputs, despite the non-linearity. It recovers the linear
performance of a network given by PCA if the decoder's weights grow correspondingly for the reconstructed input $\hat \bx$ to have the same norm as the input $\bx$.

We can find a relation between the scaling constants $\alpha^k_v$ and
$\alpha^k_w$ at long times using the ansatz \cref{eq:ansatz_longtimes} in the expression of the $\pmse$:
\begin{align}
    \begin{split}
        \pmse = \sum_{k=1}^K   \underbrace{\left\{ \tilde{\rho}_k  + \alpha_v^{k2}\EE g(\lambda^k)^2 -2 \EE \nu^k g(\lambda^k)\right\}}_{f(\alpha^k_v, \alpha^k_w)}  + \sum_{k>K}^D \tilde{\rho}_k ,
    \end{split}
\end{align}
where the second term is simply the rank $K$ PCA error. The first term
should be minimised to achieve PCA error, i.e.~$f(\alpha^k_w,
\alpha_v^{k*})\equal 0$ for all $k$. For a linear autoencoder, we find
$\alpha_v^{k*}\equal \nicefrac{1}{\alpha_w^{k}}$:  as expected, any rescaling of the encoder's weights needs to be compensated in the decoder. For a sigmoidal autoencoder instead, we find
\begin{align}
    \alpha_v^{k*} =& \frac{1}{ \alpha_w^{k}}\sqrt{ \frac{\pi}{2 } \left( 1 + \tilde{\rho}^k\alpha_w^{k2} \right)}
\end{align}
Note, that at small $\alpha^k_w$ we recover the linear scaling~$\alpha^{k*}_v~\!~\sim~\!~\left(g'(0)\alpha^k_w\right)^{-1}$.

\paragraph{The importance of untying the weights for sigmoidal autoencoders} The
need to let the encoder and decoder weights grow, resp.~shrink, to achieve PCA
error makes learning impossible in sigmoidal autoencoders with tied weights,
where $\bv^k\equal \bw^k$. Such an autoencoder evidently cannot perform the rescaling
required to enter the linear regime of its activation function, and is therefore
not able able to achieve PCA error. Even worse, the learning dynamics shown in
\cref{fig:tiedvsuntied_synthetic}(a) show that a sigmoidal autoencoder with
tied weights (blue) hardly achieves a reconstruction error better than chance,
in stark contrast to the same network with untied weights (red).

\subsection{The importance of the bias in ReLU autoencoders}
\label{sec:importance-bias-relu}
%\begin{itemize}
%    \item relu -> linear regime is positive real axis --> have to train the bias
%    \item check that without bias cannot learn & bias grows during training
%    \item error of ReLU = PCA + residual due to the bias
%\end{itemize}

Sigmoidal autoencoders achieve the PCA error by exploiting the linear region of
their activation function. We can hence expect that in order to successfully
train a ReLU AE, which is linear for all positive arguments, it is
necessary to add biases at the hidden layer. % to the encoder's weights:
% \begin{equation}
%     \hat \bx = \sum_k \bv^k g(\lambda^k + b^k).
% \end{equation}
The error curves for ReLU autoencoders shown in
\cref{fig:tiedvsuntied_synthetic}(b) show indeed that ReLU autoencoders achieve
an error close to the PCA error only if the weights are untied and the biases
are trained. If the biases are not trained, we found that a ReLU autoencoder
with $K$ hidden neurons generally achieves a reconstruction error of roughly
$K/2$, presumably because only $K/2$ nodes have a positive pre-activation and
can exploit the linear region of ReLU. Training the biases consistently resulted
in large, positive biases at the end of training, 
\cref{fig:tiedvsuntied_synthetic}(c). The large biases, however, result in a small
residual error that negatively affects the final performance of a ReLU network
when compared to the PCA performances, as can be seen from linearising the
output of the ReLU autoencoder:
\begin{equation}
    \hat{\bx} = \sum_k \bv^k \max(0, \lambda^k + b^k) = \underbrace{\sum_k \bv^k \lambda^k}_{\text{PCA}} + \underbrace{\sum_k \bv^k b^k}_{\text{residual}},
\end{equation}
where $b^k\in\reals$ is the bias of the $k$th neuron.

\subsection{Breaking the symmetry of SGD yields the exact principal components}
\label{sec:truncated-sgd}

\begin{figure*}[t!]
  \centering
  \includegraphics[width=\linewidth]{{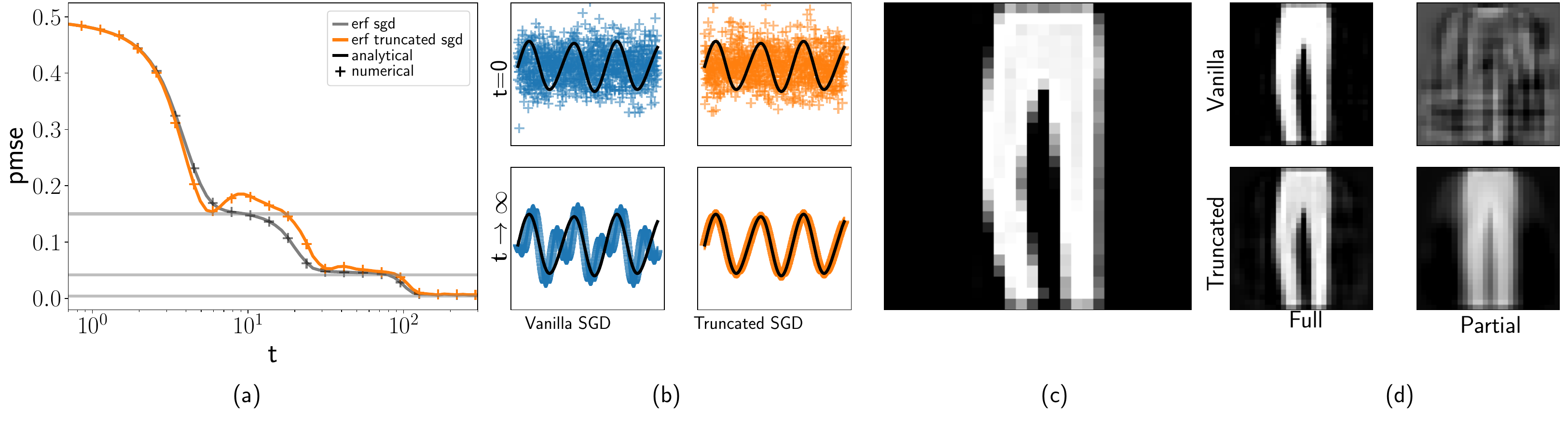}}
  \vspace*{-2em}
  \caption{\label{fig:truncated_sgd} \textbf{Breaking the symmetry
      between neurons yields the exact principal components of the data.}
    \textbf{(a)} Prediction mean-squared error~\eqref{eq:pmse} of an autoencoder
    trained using vanilla SGD (eq.~\ref{eq:sgd}, blue) and truncated SGD
    (eq.~\ref{eq:truncated_sgd}, orange). Grey horizontal lines indicate PCA
    reconstruction errors. \textbf{(b)} We show the weight vector of the first
    neuron of the autoencoder before and after training (top vs bottom) with
    vanilla and truncated SGD (left and right, resp.). In contrast to vanilla
    SGD, truncated SGD recovers the true leading principal component of the
    inputs, a sinusoidal wave (black). \textbf{(c)} Example image taken from
    Fashion MNIST. \textbf{(d)} The left column shows reconstructions of the
    image from (c) using all
    $K=64$ neurons of an autoencoder trained with vanilla (top) and truncated
    SGD (bottom) on the full Fashion MNIST database. The right column shows 
    reconstructions using only the first 5 neurons of the same
    autoencoders. \emph{Parameters:
      $\eta = 1$, $K=4$ ((a) and (b)), $K = 64$ ((c) and (d)), $bs = 1, P=60000$.} }
\end{figure*}

Linear autoencoders do not retrieve the exact principal components of the
inputs, but some rotation of them due to the rotational symmetry of the square
loss~\cite{bishop2006pattern}.  While this does not affect performance, it would
still be desirable to obtain the leading eigenvectors exactly to identify the
important features in the data.

The symmetry between neurons can be broken in a number of ways. In linear
autoencoders, previous work considered applying different amounts of
regularisation to each neuron~\cite{mianjy2018implicit, plaut2018principal, kunin2019loss, bao2020regularized} or optimising a modified
loss~\cite{oftadeh2020eliminating}. In unsupervised learning, Sanger's
rule~\cite{sanger1989optimal} breaks the symmetry by making the update of the
weights of the first neuron independent of all other neurons; the update of the
second neuron only depends on the first neuron, etc. This idea was extended to
linear autoencoders by~\textcite{oftadeh2020eliminating, bao2020regularized}. We
can easily extend this approach to non-linear autoencoders by training the
decoder weights following
% which updates the weights of a linear neuron as
% \begin{equation}
%     dv_i^k = \frac{-\eta}{D} \lambda^k (x_i -  \sum_{\ell=1}^k \lambda^\ell v^\ell_i).
% \end{equation}
% The only difference between this step and the vanilla SGD one in Eq.~\ref{eq:sgd} with $g(x)\equal x$ is that the sum only runs through a different subset of nodes for each $k$, thus breaking the symmetry. 
\begin{equation}
  \label{eq:truncated_sgd}
    \dd v_i^k = - \frac{\eta_{V}}{D} g(\lambda^k) \left(x_i  - \sum_{\ell=1}^{\textcolor{red}{k}} v^\ell_i g(\lambda^\ell) \right),
\end{equation}
where the sum now only runs up to $k$ instead of $K$. The update for the encoder
weights $\bw^k$ is unchanged from \cref{eq:sgd}, but the error term is
changed to
$\Delta_i^k\equal x_i - \sum_{\ell=1}^{\textcolor{red}{k}} v^\ell_i
g(\lambda^\ell)$.

We can appreciate the effect of this change by considering a fixed point
$\{\bW^*, \bV^*\}$ of the vanilla SGD updates \cref{eq:sgd} with $\dd\bw^{*k}\equal 0$
and $\dd \bv^{*k}\equal 0$. Multiplying this solution by an orthogonal matrix~$\bO\!\in\! O(K)$ i.e.  $\{\tilde \bW, \tilde \bV\}\equal \{\bO \bW^*, \bO \bV^*\}$
yields another fixed point, since
\begin{align}
  \label{eq:rotational_symmetry}
  \begin{split}
    \dd \tilde{v}^k_i %& = - \frac{\eta_{V}}{D} g(\tilde{\lambda}^k) \left(x_i  - \sum_{\ell=1}^k \tilde{v}^\ell_i g(\tilde{\lambda}^\ell) \right)\\
        &= - \frac{\eta_{V}}{D} g(\tilde{\lambda}^k)  \left(x_i  -  \sum_{\ell_1 \ell_2=1}^K\underbrace{\sum_{\ell=1}^K O^{\ell \ell_2} O^{\ell \ell_3}}_{\delta^{\ell_2 \ell_3}} v^{*\ell_2}_i g(\lambda^{*\ell_3} ) \right)\\
        &= - \frac{\eta_{V}}{D} g(\tilde{\lambda}^k)  \left(x_i - \sum_{\ell=1}^K v^{*\ell}_i g(\lambda^{*\ell}) \right) = 0
    \end{split}
\end{align}
for the decoder, and likewise for the encoder weights. For truncated
SGD~\eqref{eq:truncated_sgd}, the underbrace term in
\cref{eq:rotational_symmetry} is no longer the identity and the rotated weights
$\{\tilde{\bW}, \tilde{\bV}\}$ are no longer a fixed point.

We show the $\pmse$ of a sigmoidal autoencoder trained with vanilla and
truncated SGD in \cref{fig:truncated_sgd}(a), together with the theoretical
prediction from a modified set of dynamical equations. The theory captures both
learning curves perfectly, and we can see that using truncated SGD incurs no
performance penalty - both autoencoders achieve the PCA reconstruction
error. Note that the reconstruction error can increase temporarily while
training with the truncated algorithm since the updates do not minimise the
square loss directly. In this experiment, we trained the networks on a synthetic
dataset where the eigenvectors are chosen to be sinusoidal (black lines in
\cref{fig:truncated_sgd}~b). As desired, the weights of the network trained with
truncated SGD converges to the exact principal components, while vanilla SGD
only recovers a linear combination of them.

The advantage of recovering the exact principal components is illustrated with
the partial reconstructions of a test image from the FashionMNIST
database~\cite{xiao2017online}. We train autoencoders with $K\equal 64$ neurons to
reconstruct FashionMNIST images using the vanilla and truncated SGD algorithms
until convergence. We show the reconstruction using all the neurons of the
networks in the left column of \cref{fig:truncated_sgd}(d). Since the networks
achieve essentially the same performance, both reconstructions look similar. The
partial reconstruction using only the first five neurons of the networks shown
on the right are very different: while the partial reconstruction from truncated
SGD is close to the original image, the reconstruction of the vanilla
autoencoder shows traces of pants mixed with those of a sweatshirt.

\section{Representation learning on realistic data under standard conditions}
\label{sec:realistic-data}

We have derived a series of results on the training of non-linear AEs derived in
the case of synthetic, Gaussian datasets. Most of our simplifying assumptions
are not valid on real datasets, because they contain only a finite number of
samples, and because these samples are finite-dimensional and not normally distributed. Furthermore, we had to choose certain scalings of the pre-activations~\eqref{eq:local-fields} and the learning rates to ensure the existence of a well-defined asymptotic limit $D\to\infty$. Here, we show that the dynamical equations of
\cref{sec:eom} describe the dynamics of an autoencoder trained on more realistic images rather well, even if we don't perform any explicit rescalings in the simulations.

\begin{figure}[t!]
  \centering
  \includegraphics[width=\linewidth]{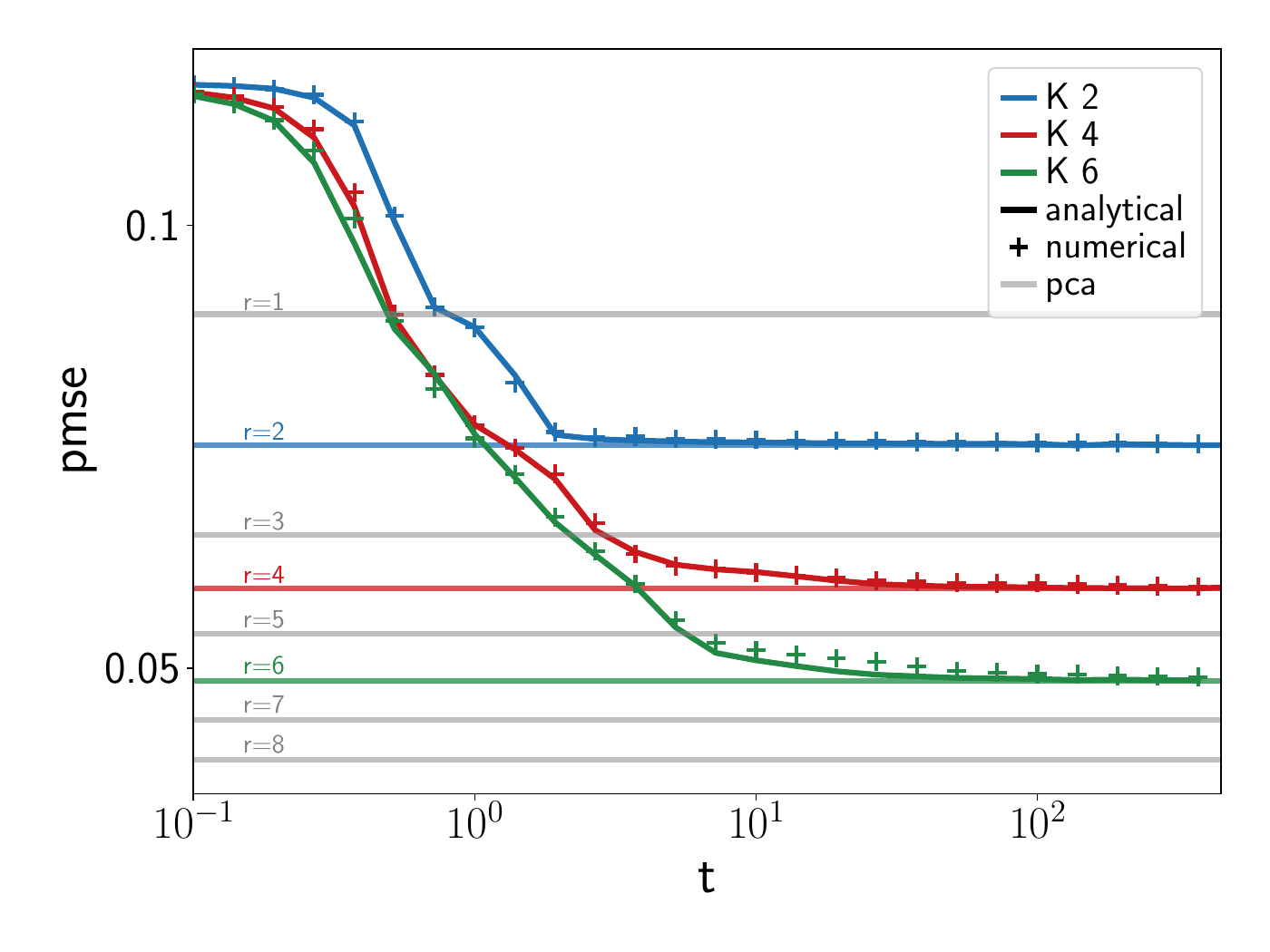}
  \vspace{-3em}
  \caption{\label{fig:realdata} \textbf{Theory vs.~simulations for sigmoidal
      autoencoders trained on CIFAR10.} Prediction mean-squared error
    $\pmse$~\eqref{eq:pmse} of a sigmoidal autoencoder with $K=2, 4, 6$ hidden
    neurons trained on 10 000 greyscale CIFAR10
    images~\cite{krizhevsky2009learning} (crosses). Solid lines show the $\pmse$
    predicted from integrating the dynamical equations of \cref{sec:eom}, using
    the empirical covariance matrix of the training inputs. Horizontal lines
    indicate the PCA error with varying
    rank $r$. \emph{Parameters:} $D=1024, \eta=1$.}
\end{figure}
\paragraph{Realistic data} In \cref{fig:realdata}, we show the $\pmse$ of autoencoders of increasing size
trained on~10k greyscale images from CIFAR10~\cite{krizhevsky2009learning}
with crosses. The solid lines are obtained from integration the equations of
motion of \cref{sec:odes}, using the empirical covariance of the training
images. The accuracy of the equations to describe the learning dynamics at all times
implies that the low-dimensional projections $\blambda$ and $\bnu$ introduced in
\cref{eq:local-fields} are very close to being normally distributed. This is by
no means obvious, since the images clearly do not follow a normal distribution
and the weights of the autoencoder are obtained from training on said images,
making them correlated with the data. Instead, \cref{fig:realdata} suggests that
from the point of view of a shallow, sigmoidal autoencoder, the inputs can be
replaced by Gaussian vectors with the same mean and covariance without changing
the dynamics of the autoencoder. In \cref{fig:cifar10vsgaussian} in the
appendix, we show that this behaviour persists also for ReLU autoencoders and,
as would be expected, for linear autoencoders. \textcite{nguyen2021analysis}
observed a similar phenomenon for tied-weight autoencoders with a number of
hidden neurons that grows polynomially with the input dimension~$D$. We also verified that several insights developed in the case of Gaussian data
carry over to real data. In particular, we demonstrate in
\cref{fig:ResultsRealData} that sigmoidal AE require untied
weights to learn, ReLU networks require adding biases to the encoder weights,
and that the principal components of realistic data are also learnt
sequentially.

% RELU: Integrating the equations with the bias becomes tedious because the pre-activations
% then have a non-zero mean. Maria did however run simulations comparing the
% dynamics of ReLU autoencoder trained on CIFAR10 vs its Gaussian clone and found
% very good agreement.

\paragraph{Relation to the Gaussian equivalence} The fact that the learning dynamics of the autoencoders trained on CIFAR10 are well captured by an equivalent Gaussian model is an example of the Gaussian equivalence principle, or Gaussian
universality, that received a lot of attention recently in the context of
supervised learning with random features~\cite{liao2018spectrum,
  seddik2019kernel, mei2021generalization} or one- and two-layer neural
networks~\cite{hu2020universality, goldt2020modelling, goldt2021gaussian,
  loureiro2021capturing}. These works showed that the performance of these different neural networks is asymptotically well captured by an appropriately chosen Gaussian model
for the data. While previous work on Gaussian equivalence in two-layer networks crucially assumes that the network has a %The result closest to our setup was obtained by \textcite{goldt2021gaussian, hu2020universality}, who showed that for two-layer networks, the Gaussian equivalence was, to a first approximation, a result of projecting high-dimensional inputs to a low-dimensional set of local fields $\blambda$ in the limit where the 
number of hidden neurons $K$ that is small compared to the input dimension, here we find that the Gaussian equivalence of
autoencoders extends to essentially any number of hidden neurons, for example
$K=D/2$, as we show in \cref{fig:cifar10vsgaussian} in the appendix.

\begin{figure}[t!]
  \centering
  \includegraphics[width=\linewidth]{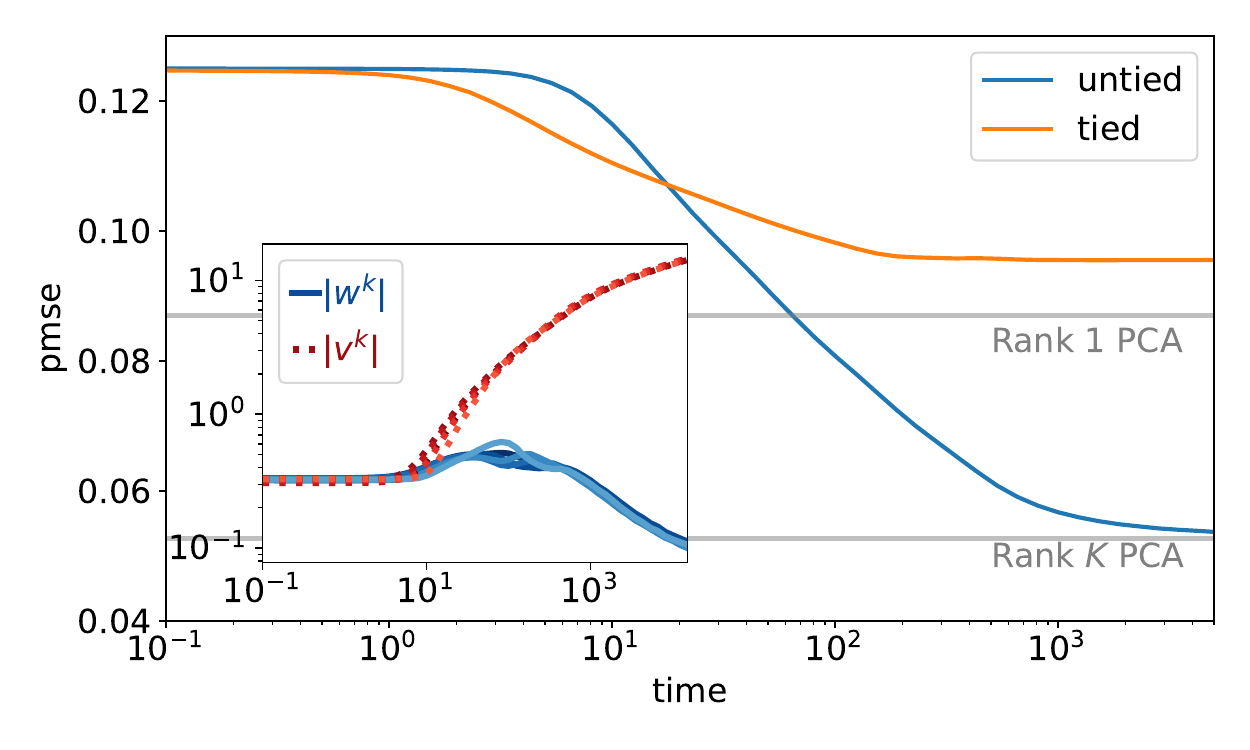}
  \caption{\label{fig:scaling-check} \textbf{Training an autoencoder without explicit rescalings} Prediction mean-squared error (pmse) of a sigmoidal autoencoder (AE) on grayscale CIFAR10 using vanilla SGD with untied (blue) and tied (orange) weights. In this experiment, we did not rescale the eigenvalues of the data, the pre-activations~\eqref{eq:local-fields}, or the learning rates. \emph{Inset}: $\ell_2$ norm of decoder (blue) and encoder (red) weights. \emph{Parameters}: $D=1024, K=5, \eta=10^{-2}$}
\end{figure}

\paragraph{Robustness of the dynamical equations to rescalings} The scaling of the local fields in \cref{eq:local-fields}, and hence of the order parameters, is necessary to ensure that in the high-dimensional limit $D \to \infty$, all three terms in the $\pmse$ are non-vanishing. As we describe in the derivation, the rescaling of the learning rates are then necessary to obtain a set of closed equations. Nevertheless, we verified that the conclusions of the paper are not an artefact of these scalings. To this end, we trained a sigmoidal AE $\hat x_i = \sum_k^K v^k_i g(w^k x)$ on grayscale CIFAR10 using SGD \emph{without} rescaling the eigenvalues of the inputs, pre-activations $\lambda^k=w^k x$, or the learning rates. As we show in \cref{fig:scaling-check}, we recover the three key phenomena described in the paper: autoencoders with tied weights fail to achieve rank-1 PCA error (orange); autoencoders with untied weights achieve rank-$K$ PCA error (blue); and for sigmoidal autoencoders, the encoder weights~\textcolor{red}{$w$} shrink, while decoder weights~\textcolor{C0}{$v$} grow to reach the linear regime (inset). These results underline that our theoretical results describe AEs under ``standard conditions''.

%% file: content/acknowledgements.tex
\section*{Acknowledgements}

We thank Galen Reeves for illuminating discussions. MR~thanks the Data Science
group at SISSA for its hospitality during a visit, where part of this
research was performed. MR acknowledges funding from the French Agence Nationale
de la Recherche under grant ANR-19-P3IA-0001 PRAIRIE.

%% file: content/appendix.tex
\begin{figure*}[t!]
  \centering
  \includegraphics[width=\linewidth]{{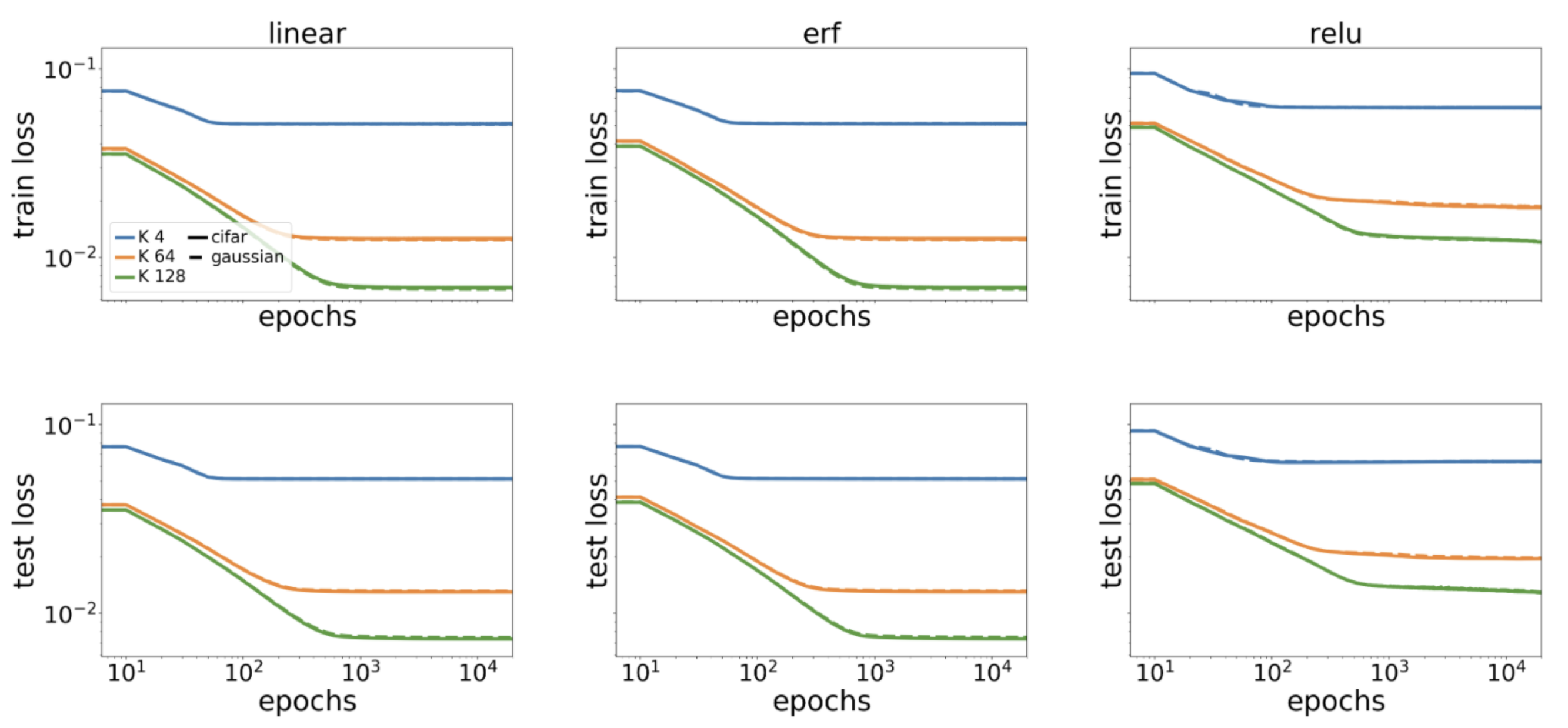}}
  \vspace*{-2em}
  \caption{\label{fig:cifar10vsgaussian}\textbf{Gaussian equivalence for
      autoencoders} Train and test mean-squared error for linear, sigmoidal
    and relu autoencoders with different numbers of hidden neurons
    ($K=4, 64, 128$ in blue, orange and green, resp.) The autoencoders were
    trained on CIFAR10 grayscale images (solid lines) or, starting from the same
    initial conditions, on Gaussian inputs with the same covariance (dashed
    lines). The agreement is essentially perfect. \emph{Parameters:} 10k
    training samples, $D=1024$, mini-batch size=1, $\eta=1$. }
\end{figure*}

\begin{figure*}[t!]
  \centering
  \includegraphics[width=\linewidth]{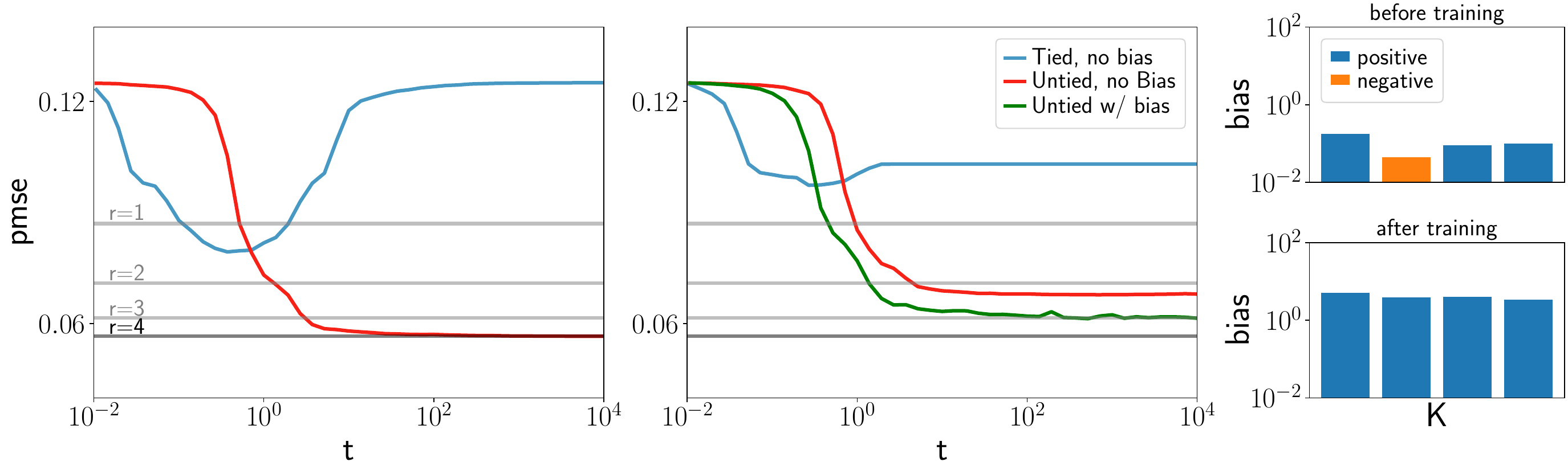}
  \vspace*{-2em}
  \caption{\label{fig:ResultsRealData}\textbf{Results derived on synthetic data
      carry over to finite real datasets} \textbf{(a)} Also in real dataset, a
    sigmoidal AE requires untied weights to attain good performances. For an
    untied student, which asymptotically performs as PCA, the dynamics slow down
    at intermediary ranks, signalling a sequential learning of different
    modes. \textbf{(b)} In order to learn, a ReLU autoencoder requires training
    the bias in the encoder's weights. The asymptotic error is given by PCA plus
    a residual due to the biases.  \emph{Parameters:} CIFAR10 and FashionMNIST
    with 10k images. batch-size=1, $\eta=1$. }
\end{figure*}

\section{Online learning algorithms for PCA}
\label{app:online-pca}

We briefly review a number of unsupervised learning algorithms for principal component analysis, leading to Sanger's rule which is the inspiration for the truncated SGD algorithm of \cref{sec:truncated-sgd}.

\paragraph{Setup} We consider inputs $\bx=(x_i), i=1,\ldots,D$ with first two moments equal to
\begin{equation}
  \EE x_i=0 \qq{and} \EE x_i x_j = \Omega_{ij}.
\end{equation}
We work in the thermodynamic limit $D\to\infty$. 

\paragraph{Hebbian learning} The Hebbian learning rule allows to obtain an estimate of the leading PC by considering a linear linear model
$y=\lambda = \sum_i\nicefrac{w_i x_i}{\sqrt D}$ with $i=1,\ldots,D$ and the loss function $\mathcal{L}(\lambda) = - \lambda^2$.
It updates the weights as:
\begin{align}
  \label{eq:hebbian}
  \dd w_i^t = -\eta_t  \nabla_{w_i} \mathcal{L}(\lambda) = \ussD \eta_t y x_i
\end{align}
In other words, the Hebbian learning rule tries to maximise the second moment of the
pre-activation $\lambda$. That using this update we obtain an estimate of the first principal component makes intuitive sense. Consider the average over the inputs of the loss:
\begin{equation}
  \EE \mathcal{L}(\lambda) = -\usD w_i \Omega_{ij} w_j.
\end{equation}
If the weight vector $\bw$ is equal to the $\tau$th eigenvector of $\Omega$, then $\EE \mathcal{L}(\lambda) = -\rho_\tau / D$, and the loss is minimised by converging to the eigenvector with the larges eigenvalue.
Also note, that similar to what we find in the main text, this result also suggests that the leading eigenvalue in the large-$D$ limit should scale as $\rho_\tau \sim D$.
The Hebbian rule has the well-known deficit that it imposes no bound on the growth of the weight vector. A natural remedy is to introduce some form
of weight decay such that
\begin{equation}
  \dd w_i^t = \frac{\eta}{\sqD} (y x_i-\kappa w_i).
\end{equation}

\noindent \textbf{Oja's rule}~\cite{oja1982simplified} offers a smart choice for
the weight decay constant $\kappa$. Consider the same linear model $y=\lambda$
trained with a Hebbian rule and weight decay $\kappa=y^2$:
\begin{align}
  \label{eq:oja}
  \dd w_i^t = \frac{\eta}{\sqD} (y x_i - y^2 w_i)
\end{align}
The purpose of this choice can be appreciated from substituting in the linear
model and setting the update rule to zero, which yields (dropping the constants)
\begin{equation}
\sum_j\Omega_{ij} w_j = \sum_{j,l}w_j \Omega_{j \ell}w_\ell w_i
\end{equation}
which is precisely the eigenvalue equation for $\Omega$.

Then, Oja's update rule is then derived from Hebb's rule by adding normalisation to the
Hebbian update~\eqref{eq:hebbian},
\begin{equation}
  \label{eq:hebb-regularised}
  w_i^{t+1} = \frac{w_i^{t} + \eta y x_i}{\left(\sum \left[w_i^{t} + \eta y x_i\right]^p\right)^{1/p}}
\end{equation}
with $p$ integer (in Oja's original paper, $p=2$.) This learning rule can be seen as a
power iteration update. Substituting for $y$, we see that the numerator corresponds to, on average, repeated multiplication of the weight
vector with the average covariance matrix. Expanding \cref{eq:hebb-regularised}
around $\eta=0$ yields Oja's algorithm (while also assuming that the weight
vector is normalised to one).
One can show that Oja's rule indeed converges to the PC with the \emph{largest}
eigenvalue by allowing the learning rate to vary with time and imposing the mild
conditions
\begin{align}
  \label{eq:eta-conditions}
  \lim_{T\to\infty} \sum_{t=0}^T \eta_t \to \infty,   \qquad \qquad \lim_{T\to\infty}  \sum_{t=0}^T \eta_t^p < \infty \qfor p > 1.
\end{align}

\paragraph{Sanger's rule~\cite{sanger1989optimal}} is also known as generalised
Hebbian learning in the literature and extends the idea behind Oja's rule to
networks with multiple outputs $y^k = w^k_i x_i$.
It allows estimation of the first few leading eigenvectors by using the update rule
is given by
\begin{equation}
  \label{eq:sanger}
  \dd w^k_i = \frac{\eta}{\sqD} y^k \left(x_i -  \sum_{\ell=1}^k y^\ell w^\ell_i\right).
\end{equation}
Note that the second sum extends only to the $k$; the dynamics of the $k$th
input vector hence only depends on weight vectors $w^\ell$ with $\ell<k$. This
dependence is one of the similarities of Sanger's rule with the Gram-Schmidt
procedure for orthogonalising a set of vectors (cf.~sec.~0.6.4 of \textcite{horn2012matrix}). The
dynamics of Sanger's rule in the online setting were studied by
\textcite{biehl1998dynamics, schloesser1999optimization}. Sanger's rule reduces
to Oja's rule for $K=1$.

\section{Online learning in autoencoders}
\label{sec:odes}
Here we derive the set of equations tracking the dynamics of shallow non-linear autoencoders trained in the one-pass limit of stochastic gradient descent (SGD). These equations are derived for Gaussian inputs $\bx\in\reals^D$ drawn from the generative model \cref{eq:dataset}, but as we discuss in \cref{sec:realistic-data}, they also capture accurately the dynamics of training on real data. 

\subsection{Statics}
The starting point of the analysis is the definition of the test error~\eqref{eq:pmse} and the identification of order parameters, i.e. low dimensional overlaps of high dimensional vectors. 
\begin{align}
    \pmse & \equiv \frac{1}{D}\sum_i\mathbb E  (x_i-\hat x_i)^2\\
    &= \frac{1}{D} \Tr \bOmega + \usD \sum_i \sum_{k, \ell} v_i^k v_i^\ell \EE  g\left(\sum_i \frac{w^k_i x_i}{\sqD} \right)g\left(\sum_i \frac{w^\ell_i x_i}{\sqD} \right) - 2 \usD \sum_i \sum_k v^k_i x_i g\left(\sum_i \frac{w^k_i x_i}{\sqD} \right).
\end{align}
First we introduce the local fields corresponding to the encoder and decoder's weights,
\begin{equation}
    \label{eq:nu}
    \lambda^k \equiv \sum_i \frac{w^k_i x_i}{\sqD} \qquad   \nu^k \equiv \sum_i \frac{v^k_i x_i}{D},
\end{equation}
as well as the order parameter tracking the overlap between the decoder weights:
\begin{equation}
  \label{eq:T0}
  T^{k\ell}_0 \equiv \usD \sum_i v^k_i v^\ell_i.
\end{equation}
Note the unusual scaling of $\nu^k$ with $D$; the intuition here is that in shallow autoencoders, the second-layer weights will be strongly correlated with the eigenvectors of the input-input covariance, and hence with the inputs, requiring the scaling $1/D$ instead of $1/\sqrt{D}$. This scaling is also the one that yields a set of self-consistent equations in the limit $D\to\infty$. The generalisation error becomes
\begin{equation}
    \pmse = \frac{1}{D} \Tr \Omega + \sum_{k, \ell} T_0^{k \ell} \EE  g(\lambda^k)g(\lambda^\ell) - 2 \sum_k \EE \nu^k g(\lambda^k).
\end{equation}
Crucially, the local fields $\lambda$ and $\nu$ are jointly Gaussian since the inputs are Gaussian. Since we have $\EE \lambda^k = \EE \nu^k = 0$, the $\pmse$ can be written in terms of the second moments of these fields only:
\begin{align}
    T_1^{k \ell} \equiv \EE \nu^k \nu^\ell = \usDD \sum_{i,j} v^k_i \Omega_{ij} v^\ell_j, \\ Q_1^{k \ell} \equiv \EE \lambda^k \lambda^\ell = \frac{1}{\sqD} \sum_{i,j} w^k_i \Omega_{ij} w^\ell_j, \\ R_1^{k \ell} \equiv \EE \nu^k \lambda^\ell = \frac{1}{D^{\nicefrac{3}{2}}} \sum_{i,j} v^k_i \Omega_{ij} w^\ell_j. \label{eq:T1}
\end{align}
The full expression of $\EE  g(\lambda^k)g(\lambda^\ell)$ and $\EE \nu^k g(\lambda^k)$ in terms of the order parameters is given, for various activation functions, in App~\ref{app:integrals}.
Note the different scalings of these overlaps with $D$. These are a direct consequence of the different scaling of the local fields $\lambda^k$ and $\nu^k$. In order to derive the equations of motion, it is useful to introduce the decomposition of $\bOmega$ in its eigenbasis:
\begin{equation}
    \Omega_{rs}=\frac{1}{D}\sum_{\tau=1}^D\Gamma_{s\tau}\Gamma_{r\tau}\rho_\tau.
\end{equation}
The eigenvectors $\Gamma_{s\tau}$ are normalised as $\sum_{i} \Gamma_{\tau i} \Gamma_{\tau' i} = D \delta_{\tau \tau'}$ and $\sum_{\tau} \Gamma_{\tau i} \Gamma_{\tau j} = D \delta_{ij}$. We further define the rotation of any $z_i \in\{w^k_i, v^k_i, x_i\}$ onto this basis as $z_\tau \equiv \nicefrac{1}{\sqD}\sum_{s=1}^D \Gamma_{\tau s} z_s$.
The order parameters are then written:
\begin{align}
    T_1^{k \ell}  = \usDD \sum_\tau \rho_\tau v_\tau^k v_\tau^\ell,\qquad
    R_1^{k \ell}   = \frac{1}{D^{3/2}} \sum_{\tau} \rho_\tau w^k_\tau v^\ell_\tau,  \qquad
    Q_1^{k \ell}   = \usD \sum_{\tau} \rho_\tau w^k_\tau w^\ell_\tau.
\end{align}

\subsection{Averages over non-linear functions of weakly correlated variables}
\label{sec:lemma1}

We briefly recall expressions to compute averages over non-linear functions of weakly correlated random variables that were also used by \textcite{goldt2020modelling}. These expressions will be extensively used in the following derivation. Consider $x$, $y\in \mathbb R$, two weakly correlated jointly Gaussian random variables with covariance matrix 
 \begin{align}
    M_{12}=
\begin{bmatrix}
    C_x &\quad \epsilon M_{12}\\
   \epsilon  M_{12} &\quad C_y
\end{bmatrix}
\end{align}

with $\epsilon\ll 1\in \mathbb R$ which encodes the weak correlation between $x$ and $y$.
Then, the expectation of the product of any two real valued functions $f,h: \mathbb{R}\to\mathbb{R}$, is given, to leading order in $\epsilon$ by:
\begin{align}
\label{eq:2pt}
    \mathop{\EE}_{(x,y)}\left[f(x)h(y) \right]=&\mathop{\EE}_{x}\left[f(x)\right]\mathop{\EE}_{y}\left[h(y)\right]\notag\\
    &+\epsilon\mathop{\EE}_{x}\left[f(x)(x-\bar x)\right]\left(C_x^{-1} M_{12} C_y^{-1}\right)\mathop{\EE}_{y}\left[h(y)(y-\bar y)\right]+O(\epsilon^2).
\end{align}
Similarly, for 3 random variables $\{x_i\}$ with $x_1$ and $x_2$ weakly correlated with $x_3$ but not between each other, i.e. $\operatorname{Cov}(x_1,x_2)=M_{12}\!\sim\!O(1)$, one can compute the expectation of product of three real valued functions $f,g , h$ to leading order in $\epsilon$ as:
\begin{align}
\begin{split}
    \EE\left[f(x_1)g(x_2)h(x_3) \right] =&\EE\left[f(x_1)g(x_2) \right]\EE\left[h(x_3) \right]\\
     &+\epsilon\frac{\EE\left[h(x_3)(x_3-\bar x_3) \right]}{ \left(C_{x_1}C_{x_2}-M_{12}^2\right) C_{x_3}}\left\{ \EE\left[f(x_1)g(x_2)(x_1-\bar x_1)\right]M_{13} C_{x_2}\right.\\
     &\hspace{11em}\left.+\EE\left[f(x_1)g(x_2)(x_2-\bar x_2)\right]M_{23}C_{x_1} \right. \\
     &\hspace{11em}\left.-\EE\left[f(x_1)g(x_2)(x_1-\bar x_1)\right]M_{12}M_{23}\right.\\
      &\hspace{11em}\left.-\EE\left[f(x_1)g(x_2)(x_2-\bar x_2)\right]M_{13}M_{12}\right\}+O(\epsilon^2).
      \label{eq:3pt}
\end{split}
\end{align}

\subsection{Derivation of dynamical equations}
\label{app:derivation_odes}
At the $\mu$th step of training, the SGD update for the rotated weights 
reads
\begin{align}
    \label{eq:sgd-rotated}
  \dd w_{\tau}^k \equiv \left(w_\tau^k\right)_{\mu+1}- \left(w_\tau^k\right)_{\mu}
  &=-\frac{\eta_{W}}{D} \frac{1}{\sqD}\sum_j^D \left(\sum_\ell^K v^\ell_j g(\lambda^\ell) - x_j\right)  v_j^k g'(\lambda^k) x_\tau - \frac{\kappa}{D} w_\tau^k\notag \\
  &=-\frac{\eta_{W}}{\sqD} \left(\sum_\ell^K T_0^{\ell k} g(\lambda^\ell) - \nu^k\right) g'(\lambda^k) x_\tau - \frac{\kappa}{D} w_\tau^k,\\
  \dd v_\tau^k &= - \frac{\eta_{V}}{D} g(\lambda^k) \left( \sum_m^K v_\tau^m g(\lambda^m) - x_\tau\right) - \frac{\kappa}{D} v_\tau^k.
\end{align}
To keep notation concise, we drop the weight decay term in the following analysis.
We can now compute the update of the various order parameters by inserting the above  into the definition of the order parameters.
The stochastic process described by the resulting equations concentrates, in the $D\to\infty$ limit, to its expectation over the input distribution. By performing the average over a fresh sample $x$, we therefore obtain a closed set of deterministic ODEs tracking the dynamics of training in the high-dimensional limit. We also show in simulations that these are able to capture well the dynamics also at finite dimension $D\sim 500$.

\paragraph{Update of $\T_0$ the overlap of the decoder's weights}
Let us start with the update equation for $T_0^{k\ell}~=~\nicefrac{1}{D} \sum_{\tau} v_\tau^k v_\tau^\ell$ which is easily obtained by using the sgd update \eqref{eq:sgd-rotated}. 
\begin{align}
    \begin{split}
        T_{0 \mu+1}^{k\ell}-T_{0 \mu}^{k\ell} = \frac{1}{D} \sum_{\tau} dv_\tau^k v_\tau^\ell + v_\tau^k dv_\tau^\ell = -\frac{\eta_V}{D}\left(\sum_a g(\lambda^k) g(\lambda^a)T_0^{a\ell} + \nu^\ell g(\lambda^k) \right) + (k\leftrightarrow \ell )
    \end{split}
\end{align}
By taking the expectation over a fresh sample $x$ and the $D\to\infty$ limit of one-pass SGD, we obtain a continuous in the normalised number of steps $s=\nicefrac{\mu}{D}$, which we interpret as a continuous time variable, as:
\begin{align}
    \begin{split}
        \frac{\partial T_0^{k\ell}}{\partial s} = -\eta_V\left(\sum_a \EE g(\lambda^k) g(\lambda^a)T_0^{a\ell} + \EE \nu^\ell g(\lambda^k) \right) + (k\leftrightarrow \ell ),
    \end{split}
\end{align}
where the notation $(k\leftrightarrow \ell )$ indicates an additional term identical to the first except for exchange of the indices $k$ and $l$.
This equation requires to evaluate $I_{22} \equiv \EE g(\lambda^k)g(\lambda^k)$, $I_{21}\equiv \EE \nu^\ell g(\lambda^k)$, which are two-dimensional Gaussian integrals with covariance matrix given by the order parameters. We give their expression in \cref{app:integrals}.
We also note that the second order term $\propto \dd v^k \dd v^\ell$ is sub-leading in the high dimensional limit we work in.

\paragraph{Update of $\T_1$ }
The update equation for $T_1^{k\ell}~=~\nicefrac{1}{D^2} \sum_{\tau}\rho_\tau v_\tau^k v_\tau^\ell$ is obtained similarly as before by using the SGD update \eqref{eq:sgd-rotated} and taking the high-dimensional limit.
\begin{align}
    \begin{split}
        \frac{\partial T_1^{k\ell}}{\partial s} =  -\eta_V \left(\sum_a \EE g(\lambda^k)g(\lambda^a)T_1^{al} + \sum_{\tau}\rho_\tau \frac{ v^\ell_\tau}{D^2}\EE g(\lambda^k)x_\tau \right) + (k\leftrightarrow \ell )
    \end{split}
\end{align}
 To make progress, we have to evaluate $\EE g(\lambda^k)x_\tau$. For this purpose it is crucial to notice that the rotated inputs $x_\tau$ are weakly correlated with the local fields:
 \begin{align}
 \begin{split}
     \EE \lambda^k x_\tau = \usD \sum_{i,j} w_i^k  \Omega_{ij} \Gamma_{\tau j} = \ussD \rho_\tau w_\tau^k, \\
     \EE \nu^k x_\tau = \frac{1}{D^{\nicefrac{3}{2}}} \sum_{i,j} v_i^k  \Omega_{ij} \Gamma_{\tau j} = \usD \rho_\tau v_\tau^k.
 \end{split}
\end{align}
Then, we can compute the expectation $ \EE g(\lambda^k) x_\tau$ using the results of \cref{sec:lemma1}, i.e. Eq.~\eqref{eq:2pt} with $f(x)= g(x)$ and $h(y)=y$, thus obtaining:
 \begin{equation}
      \EE g(\lambda^k) x_\tau = \frac{\EE \lambda^k g(\lambda^k)}{Q_1^{kk}} \frac{\rho_\tau w^k_\tau}{\sqD}
 \end{equation}
Inserting the above expression in the update of $\T_1$ yields:
\begin{align}
\frac{\partial T_1^{k\ell}}{\partial s} 
&=  \eta_V \left(\frac{\EE \lambda^k g(\lambda^k)}{Q_1^{kk}}\sum_{\tau}\rho^2_\tau \frac{v_\tau^\ell w^k_\tau}{D^{5/2}} - \sum_a \EE g(\lambda^k)g(\lambda^a)T_1^{al}  \right) + (k\leftrightarrow \ell )
 \label{eq:update_T1}
\end{align}
Notice, that in this equation we have the the appearance of $\sum_\tau \rho^2_\tau\frac{v_\tau^\ell w^k_\tau}{D^{5/2}}$, a term which cannot be simply expressed in terms of order parameters. Similar terms appear in the equation for $\Q_1$ and $\R_1$. To close the equations, we are thus led to introduce order parameter densities in the next step.

\paragraph{Order parameters as integrals over densities}
To proceed, we introduce the densities $q(\rho,s), r(\rho,s)$ and $t(\rho,s)$. These depend on $\rho$ and on the normalised number of steps $s=\nicefrac{\mu}{D}$:
\begin{align}
\begin{split}
    q^{k\ell}(\rho,s)&=\frac{1}{D \epsilon_\rho}  \sum_\tau w^k_\tau w^\ell_\tau \One,\\
    r^{k\ell}(\rho,s)&=\frac{1}{D^{\nicefrac{3}{2}} \epsilon_\rho}  \sum_\tau v^k_\tau w^\ell_\tau \One,\\
    t^{k\ell}(\rho,s)&=\frac{1}{D^2 \epsilon_\rho}  \sum_\tau v^k_\tau v^\ell_\tau \One,
\end{split}
\label{eq:op_tau}
\end{align}
where $\mathbbm{1}(.)$ is the indicator function and the limit $\epsilon_\rho\to 0$
is taken after the thermodynamic limit. 
The order parameters are obtained by integrating these densities over the spectrum of the input-input covariance matrix: 
\begin{equation}
    Q_1^{k\ell}=\frac{1}{D}\int \dd \rho  p_{\Omega}(\rho) \rho q^{k\ell}(\rho,s), \quad 
    R_1^{k\ell}=\frac{1}{D}\int \dd \rho  p_{\Omega}(\rho) \rho r^{k\ell}(\rho,s), \quad
    T_1^{k\ell}=\frac{1}{D}\int \dd \rho  p_{\Omega}(\rho) \rho t^{k\ell}(\rho,s).
\end{equation}
It follows that tracking the dynamics of the functions $q, r$ and $t$, we obtain the dynamics of the overlaps $\Q_1, \R_1$ and $\T_1$.  

\paragraph{Dynamics of $t(s, \rho)$}
The dynamics of $t$ is given straightforwardly from \cref{eq:update_T1} as:
\begin{align}
\label{eq:t_tau_ode}
 \frac{\partial t^{k\ell}(\rho, s) }{\partial s}
&= \eta_V \left( \frac{\EE \lambda^k g(\lambda^k)}{Q_1^{kk}} \tilde \rho r^{\ell k}(\rho, s) - \sum_a t^{\ell a}(\rho, s)  \EE g(\lambda^k) g(\lambda^a)
 \right)  + (k\leftrightarrow \ell )
\end{align}
where we defined the \emph{rescalled eigenvalue} $\tilde \rho \equiv \nicefrac{\rho}{D}$. Here again, straightforward algebra shows that the second order term is sub-leading in the limit of high input dimensions.

\paragraph{Dynamics of $q(s, \rho)$}
In a similar way, we compute the dynamical equation for $q(s, \rho)$ by using the encoder's weight's update in \cref{eq:sgd-rotated} and taking the expectation over the input distribution.
\begin{equation}
   \frac{\partial q^{k\ell}(\rho, s) }{\partial s} = \frac{\eta_W}{\sqD}\rho\sum_\tau \One \ w_\tau^\ell \left( \EE g'(\lambda^k) \nu^k x_\tau - \sum_a T^{ak}\EE g'(\lambda^k) g(\lambda^a)x_\tau\right) + (k\leftrightarrow \ell)
\end{equation}
In the above, we have the appearance of the expectations:
\begin{equation}
\label{eq:ABC1}
    \calA^{ka}_{\tau} = \EE [g'(\lambda^k)g(\lambda^a)x_\tau] \qquad \calB^{k}_{\tau} = \EE [g'(\lambda^k)g(\lambda^k)x_\tau] \qquad \calC^k_{\tau} = \EE [\nu^k g'(\lambda^k)x_\tau] 
\end{equation}
To compute these, we use our results for weakly correlated variables from \cref{sec:lemma1} and obtain:
\begin{align}
    \begin{split}
        \calA^{ka}_{\tau}
    =&\frac{1}{Q_1^{k k} Q_1^{a a}-\left(Q_1^{k a}\right)^{2}}\left(Q_1^{a a} \mathbb{E}\left[g^{\prime}\left(\lambda^{k}\right) \lambda^{k} g(\lambda^{a})\right]
\frac{\rho_\tau  w_\tau^k}{\sqrt D}
 -Q_1^{k a} \mathbb{E}\left[g^{\prime}\left(\lambda^{k}\right) \lambda^{a} g(\lambda^{a})\right]
\frac{\rho_\tau  w_\tau^k}{\sqrt D}
\right. \\
& \qquad \qquad  \qquad  \qquad \quad \left.-Q_1^{k a} \mathbb{E}\left[g^{\prime}\left(\lambda^{k}\right) \lambda^{k} g(\lambda^{a})\right]
\frac{\rho_\tau  w_\tau^a}{\sqrt D}
+Q_1^{k k} \mathbb{E}\left[g^{\prime}\left(\lambda^{k}\right) \lambda^{a} g(\lambda^{a})\right]
\frac{\rho_\tau  w_\tau^a}{\sqrt D}
\right)\\
    \calB^{k}_{\tau}=&\frac{\mathbb{E}\left[g^{\prime}\left(\lambda^{k}\right) \lambda^{k} g(\lambda^{k})\right]}{Q_1^{k k}}  \frac{\rho_\tau  w_\tau^k}{\sqrt D}\\
\calC^k_{\tau} =&\frac{1}{Q_1^{k k} T_1^{k k}-\left(R_1^{k k}\right)^{2}}\left(T_1^{k k} \mathbb{E}\left[g^{\prime}\left(\lambda^{k}\right) \lambda^{k} \nu^k\right]
\frac{\rho_\tau  w_\tau^k}{\sqrt D}
 -R_1^{k k} \mathbb{E}\left[g^{\prime}\left(\lambda^{k}\right) \nu^{k 2}\right]
\frac{\rho_\tau  w_\tau^k}{\sqrt D}
\right. \\
& \qquad \qquad  \qquad  \qquad \quad \left.-R_1^{k k} \mathbb{E}\left[g^{\prime}\left(\lambda^{k}\right) \lambda^{k} \nu^k\right]
\frac{\rho_\tau  v_\tau^k}{ D}
+Q_1^{k k} \mathbb{E}\left[g^{\prime}\left(\lambda^{k}\right)  \nu^{k 2}\right]
\frac{\rho_\tau  v_\tau^k}{ D}
\right)
    \end{split}
\end{align}
Using these results in the equations for $q(\rho, s)$ we find:
\begin{align}
\label{eq:q_tau_ode}
    \begin{split}
 \frac{\partial q^{k\ell}(\rho, s) }{\partial s}
  &= \eta_W D {\tilde \rho} \left\{
    \frac{1}{Q_1^{k k} T_1^{k k}-\left(R_1^{k k}\right)^{2}}\left(T_1^{k k} \mathbb{E}\left[g^{\prime}\left(\lambda^{k}\right) \lambda^{k} \nu^k\right]
q^{kl}(\rho, t)
 -R_1^{k k} \mathbb{E}\left[g^{\prime}\left(\lambda^{k}\right) \nu^{k 2}\right]
q^{kl}(\rho, t)
\right.\right. \\
& \qquad \qquad  \qquad  \qquad \qquad \qquad \quad  \left.\left.-R_1^{k k} \mathbb{E}\left[g^{\prime}\left(\lambda^{k}\right) \lambda^{k} \nu^k\right]
r(\rho, t)^{kl}
+Q_1^{k k} \mathbb{E}\left[g^{\prime}\left(\lambda^{k}\right)  \nu^{k 2}\right]
r(\rho)^{kl}
\right) 
    \right. \\
    &\qquad \quad -\sum_{a\neq k}T_0^{ka}\left.\frac{1}{Q_1^{k k} Q_1^{a a}-\left(Q_1^{k a}\right)^{2}}  
    \left(Q_1^{a a} \mathbb{E}\left[g^{\prime}\left(\lambda^{k}\right) \lambda^{k} g(\lambda^{a})\right]
q^{kl}(\rho, t)
 -Q_1^{k a} \mathbb{E}\left[g^{\prime}\left(\lambda^{k}\right) \lambda^{a} g(\lambda^{a})\right]
q^{kl}(\rho, t) \right.\right.\\
& \qquad \qquad\qquad \quad   \qquad  \qquad \qquad \qquad \left.\left.-Q_1^{k a} \mathbb{E}\left[g^{\prime}\left(\lambda^{k}\right) \lambda^{k} g(\lambda^{a})\right]
q(\rho)^{al}
+Q_1^{k k} \mathbb{E}\left[g^{\prime}\left(\lambda^{k}\right) \lambda^{a} g(\lambda^{a})\right]
q(\rho, t)^{al}
\right)\right.\\
&\qquad \quad -T_0^{kk}\left. \frac{\mathbb{E}\left[g^{\prime}\left(\lambda^{k}\right) \lambda^{k} g(\lambda^{k})\right]}{Q_1^{k k}}  q^{kl}(\rho, t) \right\}\\
&+ \left(l\leftrightarrow k\right)
    \end{split}
\end{align}
Note, that as explained in the main text, in order to find a well defined $D\to\infty$ limit, we rescale the learning rate of the encoder's weights as $\eta_W = \nicefrac{\eta}{D}$.

\paragraph{Dynamics of $r(s, \rho)$}
Lastly, we obtain the equation for $r(\rho, s)$ in the same way as the two previous ones. We use \cref{eq:sgd-rotated} and evaluate the expectation over a fresh sample $x$. 
\begin{align}
\label{eq:r_tau_ode}
\begin{split}
      \frac{\partial r^{k\ell}(\rho, s) }{\partial s} &= \eta \left(\frac{1}{\epsilon_g D^{\nicefrac{3}{2}}} \sum_\tau \One w^\ell_\tau \EE x_\tau g(\lambda^k)
   - \sum_a \frac{1}{\epsilon_g D^{\nicefrac{3}{2}}} \sum_\tau \One w^\ell_\tau v^a_\tau  \EE g(\lambda^k) g(\lambda^a) \right)\\
   &+ \eta \left( \frac{1}{\epsilon_g D} \sum_\tau \One  v^k_\tau 
         \EE g'(\lambda^\ell) \nu^\ell x_\tau - \sum_a T_0^{al} \frac{1}{\epsilon_g D} \sum_\tau \One  v^k_\tau \EE g'(\lambda^\ell) g(\lambda^a) x_\tau
        \right)
\end{split}
\end{align}
We can evaluate the expectations as before (\ref{sec:lemma1}). Thus, we obtain the equation of motion for $r(\rho, s)$ as:
\begin{align}
    \begin{split}
       \frac{\partial r^{k
       \ell}(\rho,s)}{\partial s}  =  
    &\eta \tilde \rho \left\{
    \frac{1}{Q_1^{k k} T_1^{k k}-\left(R_1^{k k}\right)^{2}}\left(T_1^{k k} \mathbb{E}\left[g^{\prime}\left(\lambda^{k}\right) \lambda^{k} \nu^k\right]
r^{\ell k}(\rho,t)
 -R_1^{k k} \mathbb{E}\left[g^{\prime}\left(\lambda^{k}\right) \nu^{k 2}\right]
r^{\ell k}(\rho,t)
\right.\right. \\
& \qquad \qquad  \qquad  \qquad \quad \left.\left.-R_1^{k k} \mathbb{E}\left[g^{\prime}\left(\lambda^{k}\right) \lambda^{k} \nu^k\right]
t^{k\ell}(\rho,t)
+Q_1^{k k} \mathbb{E}\left[g^{\prime}\left(\lambda^{k}\right)  \nu^{k 2}\right]
t^{k\ell}(\rho,t)
\right) 
    \right. \\
    &-\sum_{a\neq k}T_0^{ka}\left.\frac{1}{Q_1^{k k} Q_1^{a a}-\left(Q_1^{k a}\right)^{2}}  
    \left(Q_1^{a a} \mathbb{E}\left[g^{\prime}\left(\lambda^{k}\right) \lambda^{k} g(\lambda^{a})\right]
r^{lk}(\rho,t)
 -Q_1^{k a} \mathbb{E}\left[g^{\prime}\left(\lambda^{k}\right) \lambda^{a} g(\lambda^{a})\right]
r^{lk}(\rho,t) \right.\right.\\
& \qquad \qquad  \qquad  \qquad \quad \left.\left.-Q_1^{k a} \mathbb{E}\left[g^{\prime}\left(\lambda^{k}\right) \lambda^{k} g(\lambda^{a})\right]
r^{la}(\rho,t)
+Q_1^{k k} \mathbb{E}\left[g^{\prime}\left(\lambda^{k}\right) \lambda^{a} g(\lambda^{a})\right]
r^{la}(\rho,t)
\right)\right.\\
&-T_0^{kk}\left. \frac{\mathbb{E}\left[g^{\prime}\left(\lambda^{k}\right) \lambda^{k} g(\lambda^{k})\right]}{Q_1^{k k}}  r^{\ell k}(\rho,t) \right\}\\
&+ \eta \left\{\frac{\EE \lambda^\ell g(\lambda^\ell)}{Q_1^{\ell \ell}}{\tilde \rho}q(\rho,t) - \sum_a r^{a k}(\rho,t) \EE g(\lambda^\ell)g(\lambda^a) \right\}
    \end{split}
\end{align}

\subsection{Simplification of the equations for spiked covariance matrices}
In the case of the synthetic dataset defined as $\bx =\bA \bc + \bxi$, the spectrum of the covariance matrix is decomposed into a \emph{bulk} of small, i.e. $\order{1}$ eigenvalues, of continuous support,  and a \emph{few} outlier eigenvalues taking values of order $\order{D}$ (see \cref{fig:figure1}). 
This allows to obtain $M$ equations for controlling the evolution of the spikes modes and one equation controlling the evolution of the bulk for all order parameters.
Indeed we can simplify the equations of motion of the bulk eigenvalues (i.e. those for which $\rho \sim O(1)$) by neglecting terms proportional to $\nicefrac{\rho}{D}\ll1$. 
Doing so leads to:
\begin{align}
\label{eq:bulk_modes_ode}
    \begin{split}
     \frac{\partial t^{k\ell}(\rho, s) }{\partial s}
&= - \eta  \sum_a t^{\ell k}(\rho, s)  \EE g(\lambda^k) g(\lambda^a)  + (k\leftrightarrow \ell )-\kappa t^{k\ell}(\rho, s)\\
         \frac{\partial r^{kl}(\rho,s)}{\partial s} &= -\eta  \sum_a \EE g(\lambda^k) g(\lambda^a) r^{a k}(\rho,t)-\kappa r^{k\ell}(\rho, s)\\
         \frac{\partial q^{k\ell}(\rho, s) }{\partial s}
  &=-\kappa q^{k\ell}(\rho, s).
    \end{split}
\end{align}
We can define \emph{bulk} order parameters, that take into account the contribution from the bulk modes as:
\begin{align}
    \begin{split}
        \T_{\bulk} = \int_{\rho\in {\bulk}}\dd\!p_{\Omega}(\rho)\frac{\rho}{D} t(\rho),\qquad 
        \R_{\bulk} = \int_{\rho\in {\bulk}}\dd\!p_{\Omega}(\rho)\frac{\rho}{D} r(\rho),\qquad
        \Q_{\bulk} = \int_{\rho\in {\bulk}}\dd\!p_{\Omega}(\rho)\frac{\rho}{D} q(\rho).
    \end{split}
\end{align}
Note that even though the integrals involve $\nicefrac{\rho}{D}\sim O(\nicefrac{1}{D})$ since we are integrating over a large number ($O(D)$) of modes, the result is of order 1. 
The equation for these overlaps are thus obtained as the integrated form of Eq.~\eqref{eq:bulk_modes_ode}:
\begin{align}
\label{eq:Bulk_Odes}
    \begin{split}
     \frac{\partial T_{\bulk}^{k\ell} }{\partial s}
=& - \eta  \sum_a T_{\bulk}^{\ell a}  \EE g(\lambda^k) g(\lambda^a)  + (k\leftrightarrow \ell )-\kappa T_{\bulk}^{k\ell}\\
        \frac{\partial  R_{\bulk}^{k\ell}}{\partial s} =& -\eta \sum_a R_{\bulk}^{a k}(\rho,t) \EE g(\lambda^k) g(\lambda^a)-\kappa R^{k\ell}_{\bulk}\\
      \frac{\partial Q_{\bulk}^{k\ell}}{\partial s}
  =&-\kappa Q_{\bulk}^{k\ell}.
    \end{split}
\end{align}
The order parameters can be decomposed into the contribution from the bulk eigenvalues and those of the spikes as:
\begin{align}
    \begin{split}
    \T_1 = \sum_{i\in\text{outliers}}^M \tilde{\rho}_i t_i + T_{\bulk}
    \end{split}
\end{align}
and similarly for $\Q_1$ and $\R_1$. The $M$ spike modes $t_i, r_i$ and $q_i$ obey the full equations Eq.~\eqref{eq:t_tau_ode} , Eq.~\eqref{eq:r_tau_ode} and Eq.~\eqref{eq:q_tau_ode}. In particular, it is clear from Eq.~\eqref{eq:Bulk_Odes} that for any non-zero weight decay constant $\kappa$, the bulk contribution to $\Q_1$ will decay to $0$ in a characteristic time $\kappa^{-1}$. Further note that the equations for $\T_{\bulk}$ and $\R_{\bulk}$ result in an exponential decay of the bulk modes towards $0$. 

\section{Reduced equations for long-time dynamics of learning} 
\label{sec:app:reduced-eom}

In this section we illustrate how to use the equations of motion in order to study the long training time dynamics of shallow non-linear autoencoders. At sufficiently long times, we have seen that the weights of the network span the leading PC subspace of the covariance matrix. We restrict to the case in which the eigenvectors are recovered directly. The case in which a rotation of them is found follows straightforwardly. We also restrict to the matched case in which $K=M$.% so that the bulk modes will be $0$ at long times.

Consider the following ansatz on the weights configuration: $w_i^k = \nicefrac{\alpha_\omega}{\sqD}\Gamma_i^k$ and $v_i^k = \alpha^k_v\Gamma_i^k$. We defined dynamical constants $\alpha^k_w, \alpha_v^k\in \reals$ which control the norm of the weights. Using this ansatz, the order parameters take the form:
\begin{align}
\label{eq:egein-op}
    Q_1^{k\ell}= \alpha^{k2}_w \tilde\rho_k \delta^{k\ell}\qquad
    R_1^{k\ell}= \alpha^{k}_w  \alpha^k_v \tilde\rho_k \delta^{k\ell}\qquad
    T_1^{k\ell}= \alpha^{k2}_v \tilde\rho_k \delta^{k\ell}\qquad
    T_0^{k\ell}= \alpha^{k2}_v \delta^{k\ell}
\end{align}
Using the above, we can rewrite the $\pmse$ as:
\begin{equation}
    \pmse(\alpha^{k}_w,  \alpha^k_v) = \sum_k\left\{
    T_0^{kk}\EE g(\lambda^k)^2 - 2\EE \nu^k g(\lambda^k) + \tilde  \rho_k
    \right\} + \sum_{i>K} \tilde\rho_i
\end{equation}
The first observation is that the generalisation error reaches a minimum when the term in bracket $f^k$ is minimised. Minimising it leads to an equation $\alpha^{*k}_v(\alpha_w)$ at which $f^k(\alpha^{*k}_v(\alpha_w),\alpha_w)=0$ and the $\pmse$ is equal to the PCA reconstruction error. 

For a linear activation function, we have \begin{equation}
    \pmse(\alpha^{k}_w,  \alpha^k_v) = \sum_k\underbrace{\left\{
    \alpha^{k2}_v\alpha^{k2}_w\tilde \rho_k 
    -  2 \alpha^{k}_w \alpha^k_v \tilde \rho_k + \tilde  \rho_k
    \right\}}_{ f_k(\alpha^k_v, \alpha^{k}_w)} + \underbrace{\sum_{i>K} \tilde\rho_i}_{\text{PCA reconstruction error}}
\end{equation}
The $\pmse$ is a function only of the product $\alpha^{k 2} \equiv \alpha_w^k \alpha^k_v$ and is minimal and equal to the PCA reconstruction error whenever: 
\begin{equation}
    \frac{\partial f_k(\alpha_v^k, \alpha^k_w)}{\partial\alpha_v^l}=0 \Longrightarrow \alpha_v^{*k} = \frac{1}{\alpha^k_w},\quad \forall k \qquad  \text{and}   \qquad f_k(\frac{1}{\alpha^k_w}, \alpha^k_w) = 0
\end{equation}
The above is nothing but the intuitive result that for a linear autoencoder, any rescaling of the encoder's weights can be compensated by the decoder's weights. 

For a sigmoidal autoencoder, instead, the expressions of the integrals given in App.~\ref{app:integrals}, give:

\begin{equation}
    \pmse(\alpha^{k}_w,  \alpha^k_v) = \sum_k\underbrace{\left\{\frac{2}{\pi}
    \alpha^{k2}_v\frac{\alpha^{k2}_w\tilde \rho_k }{1+\alpha^{k2}_w\tilde \rho_k} 
    - \sqrt{\frac{2}{\pi (1+\alpha^{k2}_w\tilde \rho_k)}} 2 \alpha^{k}_w \alpha^k_v \tilde \rho_k + \tilde  \rho_k
    \right\}}_{ f_k(\alpha^k_v, \alpha^{k}_w)} + \sum_{i>K} \tilde\rho_i
\end{equation}
Differentiating $f_K$ with respect to $\alpha^{k}_w$, and requiring the derivative to be $0$, we find the equation:
\begin{align}
\label{eq:solution_erf_w}
\begin{split}
        2 \pi \left(1 + \tilde \rho^k \alpha_\omega^{k2} \right)^2 =& 4  \alpha_\omega^{k2}  \alpha_v^{k2} \left(1 + \tilde \rho^k \alpha_\omega^{k2} \right)\\
        \Longrightarrow  
    \alpha_v^{k*2}(\alpha_w) =& \frac{\pi}{2}\left(\frac{1}{\alpha_w^{2k}}+\tilde \rho\right)
\end{split}
\end{align}
We point out that this solution is independent of the sign of $\alpha^k_\omega$ or $\alpha^k_v$ as recovering the eigenvectors or minus the eigenvectors is equivalent.
Replacing this solution for $\alpha_\omega^k$ in the expression for the $\pmse$, we find $f^k(\alpha_v^{k*}(\alpha_w), \alpha_w^k)=0$. The autoencoder reaches the same reconstruction error as the one achieved by PCA.

\paragraph{Ansatz in the equations of motion}
The ansatz Eq.~\ref{eq:ansatz_longtimes} results in order parameters of the form: 
\begin{equation}
    q^{kl}(\rho_m) = \frac{\alpha^{k2}_w}{D}\delta^{kl}\delta^{k}_m \qquad
    r^{kl}(\rho_m) = \frac{\alpha^{k}_w \alpha^k_v}{D}\delta^{kl}\delta^{k}_m \qquad
    t^{kl}(\rho_m) = \frac{\alpha^{k2}_v}{D}\delta^{kl}\delta^{k}_m 
\end{equation}

Inserting these expressions into the full dynamical equations of motion Eq.~\eqref{eq:q_tau_ode}, Eq.~\eqref{eq:r_tau_ode} and Eq.~\eqref{eq:t_tau_ode}, allows us to finnd a simplified set of $2 K$ ODEs for the scaling constants $\alpha^k_v$ and $\alpha^k_w$:
\begin{align}
  \label{eq:reduced-eom}
    \begin{split}
    \frac{\partial \alpha_v^k}{\partial s} &= \frac{\tilde{\rho}^k \alpha_w^k}{\pi\left(\alpha_w^{k2}  \tilde{\rho} +1 \right)}\sqrt{2 \pi \left( \alpha_w^{k2}  \tilde{\rho} +1 \right) - 2 \alpha_v^k \alpha_w^k}\\
     \frac{\partial \alpha_w^k}{\partial s} &= \alpha^k_v \tilde{\rho}^k\sqrt{\frac{2}{\pi\left( \alpha_w^{k2}  \tilde{\rho} +1 \right)^{3}}} \left(1 - \sqrt{\frac{2\left( \alpha_w^{k2}  \tilde{\rho} +1 \right)}{\pi \left(2 \alpha_w^{k2}  \tilde{\rho} +1 \right)}} \alpha_v^k \alpha_w^k   \right)
    \end{split}
\end{align}
The validity of the equations is verified in Fig.~\ref{fig:longtimedynamics} where we compare the result of integrating them to simulations. We provide a Mathematica notebook which allows to derive these equations on the \href{https://github.com/anon}{Github} repository of this paper.

\begin{figure}[t!]
  \centering
  \includegraphics[width=0.8\linewidth]{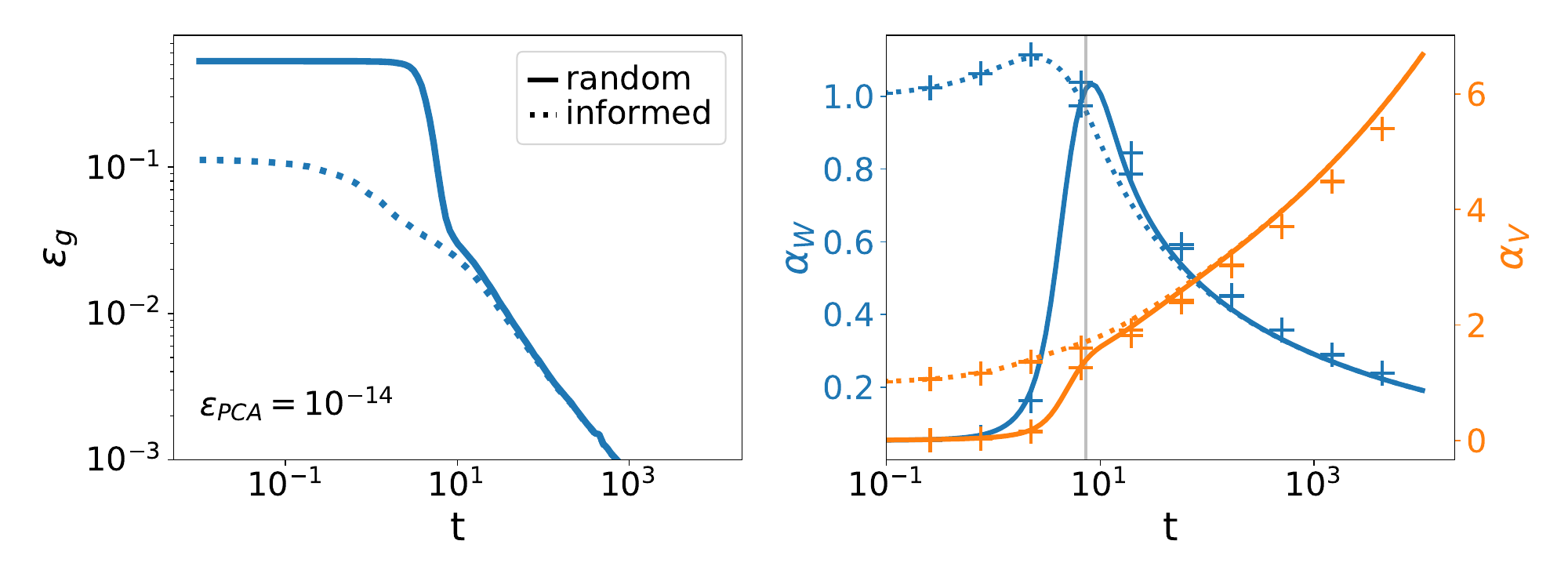}
  \vspace{-0.5cm}
  \caption{\label{fig:app:shrink_and_grow} \textbf{Learning in non-linear shallow AE occurs in two phases} \textbf{(a)} The $\pmse$ of an auto-encoder trained from random initial conditions first decays exponentially and then as a power law in time. \textbf{Dynamics of the scaling constants $\alpha_v$ and $\alpha_w$} The reduced set of $2K$ ODEs match the results of simulations. The dynamics of an AE initialised randomly  with random initial conditions become indistinguishable with those of an AE whose weights are initialised proportional to the PCs. This shows that the network first recovers the leading PCs subspace and then adjusts the norm of the weights in order to recover the linear regime.    \emph{Parameters: $\eta = 1, K = 64, bs = 1$}
    }
\end{figure}

To validate our calculations for autoencoders with finite $D$, we  trained an autoencoder from three distinct initial conditions: A \emph{random} one, an \emph{informed} configuration in which the weights are initialised as in Eq.~\ref{eq:ansatz_longtimes} with $\alpha_v^k = \alpha_w^k = 1$ and a \emph{perfect} one, in which additionally, $\alpha_v^k$ is initialised from the minimal $\pmse$ solution Eq.~\ref{eq:solution_erf_w}. The first observation is that the $\pmse$ of the network started with \emph{perfect} initial conditions remains at the PCA reconstruction error, thus validating Eq.~\ref{eq:solution_erf_w}. Since we have a noiseless dataset, the PCA error is given by the numerical error, however we note that the same conclusion caries over to noisy datasets with finite PCA reconstruction error. In \cref{fig:app:shrink_and_grow}, we plot the $\pmse$ of the randomly  initialised and informed autoencoders on the left as well as the evolution of $\alpha_v^k$ and $\alpha_w^k$, i.e. of the overlap between the weights and the eigenvectors during training, on the right. We can see that the network with random initial weights learns in two phases: first the $\pmse$ decays exponentially until it reaches the error of the aligned network. Then, its $\pmse$ coincides with that of the informed network and the error decays as a power-law. This shows that during the first phase of exponential learning, the network recovers the leading PC subspace. This occurs early on in training. In the second phase, the networks adjust the norm of the weights to reach PCA performances. As discussed in the main text, this is achieved in the sigmoidal network by shrinking the encoder's weights, so as to recover the linear regime of the activation function. It keeps the norm of the reconstruction similar to the one of the input by and growing the decoder's weights.

% \subsection{Robustness of results to scaling choices}

\section{Analytical formula of the integral}
\label{app:integrals}
In this subsection we define $C=(c_{ij})_{i,j=1, ..,k}$ the covariance matrix of the variables $\{x_i\}_{i=1,..,k}$ appearing in the expectation. For e.g. $\EE x_1x_2 - \EE x_1\EE x_2 = c_{12}$.  
\paragraph{Sigmoidal activation function}
\begin{align}
        I_2 &\equiv  \EE g(x_1) g(x_2)  = \frac{2}{\pi}\left[\frac{c_{12}}{\sqrt{1+c_{11}}\sqrt{1+c_{22}}}\right]\\
        J_2 &\equiv  \EE x_1 g(x_2)   =  \sqrt{\frac{2}{\pi(1+c_{22})}} c_{12}\\
        I_3&\equiv  \EE g'(x_1)x_2 g(x_3)  = \frac{2}{\pi}\frac{1}{\sqrt{ (1+c_{11})(1+c_{33})-c_{13}^2 }}\frac{c_{23}(1+c_{11})-c_{12}c_{13}}{1+c_{11}}\\
        I_{21} &\equiv  \EE g'(x_1) x_1 x_2   = \sqrt{\frac{2}{\pi }}\frac{ c_{12}}{(c_{11}+1)^{3/2}}\\
        I_{22} &\equiv  \EE g'(x_1) x_2^2   = \sqrt{\frac{2}{\pi }} \frac{ \left(c_{11} c_{22}-c_{12}^2+c_{22}\right)}{(c_{11}+1)^{3/2}}
\end{align}
\paragraph{Linear activation function}
\begin{align}
        &I_2 = J_2 = I_{21} \equiv  \EE x_1 x_2  = c_{12}\\
         & I_{22} \equiv  \EE x_2^2  = c_{22}\\
         &  I_{3} \equiv  \EE x_2 x_3  = c_{23}\\
\end{align}

\paragraph{ReLU activation function}
\begin{align}
        I_2 &\equiv  \EE g(x_1) g(x_2)  =\frac {1} {8\pi}\left[
   2 \sqrt {c_ {11} c_ {22} - c_ {12}^2} + c_ {12}\pi + 
    2 c_ {12}+ \arctan{\frac {c_ {12}} {\sqrt {c_ {11} c_ {22} - 
           c_ {12}^2}}} \right] \\
        J_2 &\equiv  \EE x_1 g(x_2)   = \frac{c_{12}}{2}
\end{align}